\documentclass[11pt]{article}

\usepackage[preprint]{acl}
\usepackage{multirow}
\usepackage{times}
\usepackage{latexsym}
\usepackage{textcomp}
\usepackage{amsmath}
\usepackage{amssymb}
\usepackage{tcolorbox}
\usepackage{svg}
\usepackage[normalem]{ulem}
\useunder{\uline}{\ul}{}
\usepackage{enumitem}
\usepackage{amsthm}
\usepackage{subcaption}
\newtheorem{theorem}{Theorem}[section]

\usepackage[T1]{fontenc}

\usepackage[utf8]{inputenc}

\usepackage{microtype}

\usepackage{inconsolata}
\usepackage{array}
\usepackage{graphicx}
\usepackage{booktabs}
\title{Beyond Binary: Turning Partial Success into Dense Verifiable Rewards \\ for Reinforcement Learning in Code Generation}

\newcommand*\samethanks[1][\value{footnote}]{\footnotemark[#1]}
\author{
 \textbf{Longwen Wang\textsuperscript{1,3}}\thanks{Equal contribution.}\thanks{Work done during internship at TeleAI.},
 \textbf{Yirui Liu\textsuperscript{1}}\samethanks[1], 
 \textbf{Xuan'er Wu\textsuperscript{1}},
 \textbf{Xiaohui Hu\textsuperscript{2}},
 \textbf{Yuankai Fan\textsuperscript{1}},
 \\
 \textbf{Kaidong Yu\textsuperscript{2}},
 \textbf{Qizhen Weng\textsuperscript{1}},
 \textbf{Wei Xi\textsuperscript{2}}\thanks{Corresponding authors},
 \textbf{Xuelong Li\textsuperscript{1}}\samethanks
\\
\\
 \textsuperscript{1}Institute of Artificial Intelligence, China Telecom (TeleAI)
 \\
 \textsuperscript{2}Xingchen AGI Lab, China Telecom Artificial Intelligence Technology (Beijing) Co., Ltd
 \\
 \textsuperscript{3}National Key Laboratory of Human-Machine Hybrid Augmented Intelligence, Xi'an Jiaotong University
\\
 {
  \tt\small longwenwang12@gmail.com, xiwei@xjtu.edu.cn, xuelong\_li@ieee.org
 }
}

\begin{document}
\maketitle
\begin{abstract}
Effective reward design is a central challenge in Reinforcement Learning (RL) for code generation. Mainstream test-suite-level outcome rewards enforce functional correctness but induce sparsity, while external Reward Models (RMs) provide dense supervision at the cost of misalignment and additional overhead.  Since code evaluation naturally yields multiple test-case-level outcomes, partial success—passing a subset of test cases—offers an intrinsic, verifiable source of dense supervision. In this paper, we propose \textbf{VeRPO} (\textbf{V}erifiable D\textbf{e}nse \textbf{R}eward \textbf{P}olicy \textbf{O}ptimization), an RL framework that systematically turns verifiable partial success into reliable dense rewards. We analyze partial-success rewards using a weighted sum formulation, theoretically identifying a critical cardinality bias that causes policy updates to disproportionately favor gains from easy-test successes over progress on frontier tests.
Based on this, VeRPO introduces a dynamic, density-calibrated local reward that explicitly corrects this bias and provides robust dense supervision from partial success. To enhance alignment with end-to-end functional correctness, VeRPO further integrates the local dense reward with global execution outcomes. Extensive experiments across diverse benchmarks and settings demonstrate that VeRPO outperforms outcome-driven and RM-based baselines, achieving up to +8.83 pass@1 gain with negligible time cost (< 0.02\%) and zero GPU memory overhead.
\end{abstract}

\section{Introduction}\label{Introduction}
Code generation has emerged as a core capability of Large Language Models (LLMs), empowering LLMs to address competitive programming tasks through single-turn solution synthesis \cite{roziere2023code-single-turn-code,wang2021codet5-single-turn} or iterative multi-turn refinement \cite{dong2025survey-multi-turn-code,zheng2024opencodeinterpreter-multi-turn-code}, where each generated executable solution is evaluated against a suite of problem-specific test cases. Within this domain, Reinforcement Learning (RL) has significantly scaled performance, with its effectiveness hinging on the design of reward signals \cite{mroueh2025reinforcement-rlvrwithcode, chen2025R1-Code-Interpreter, lambert2024tulu-RLVR}.

\begin{figure}[t]
  \includegraphics[width=\columnwidth]{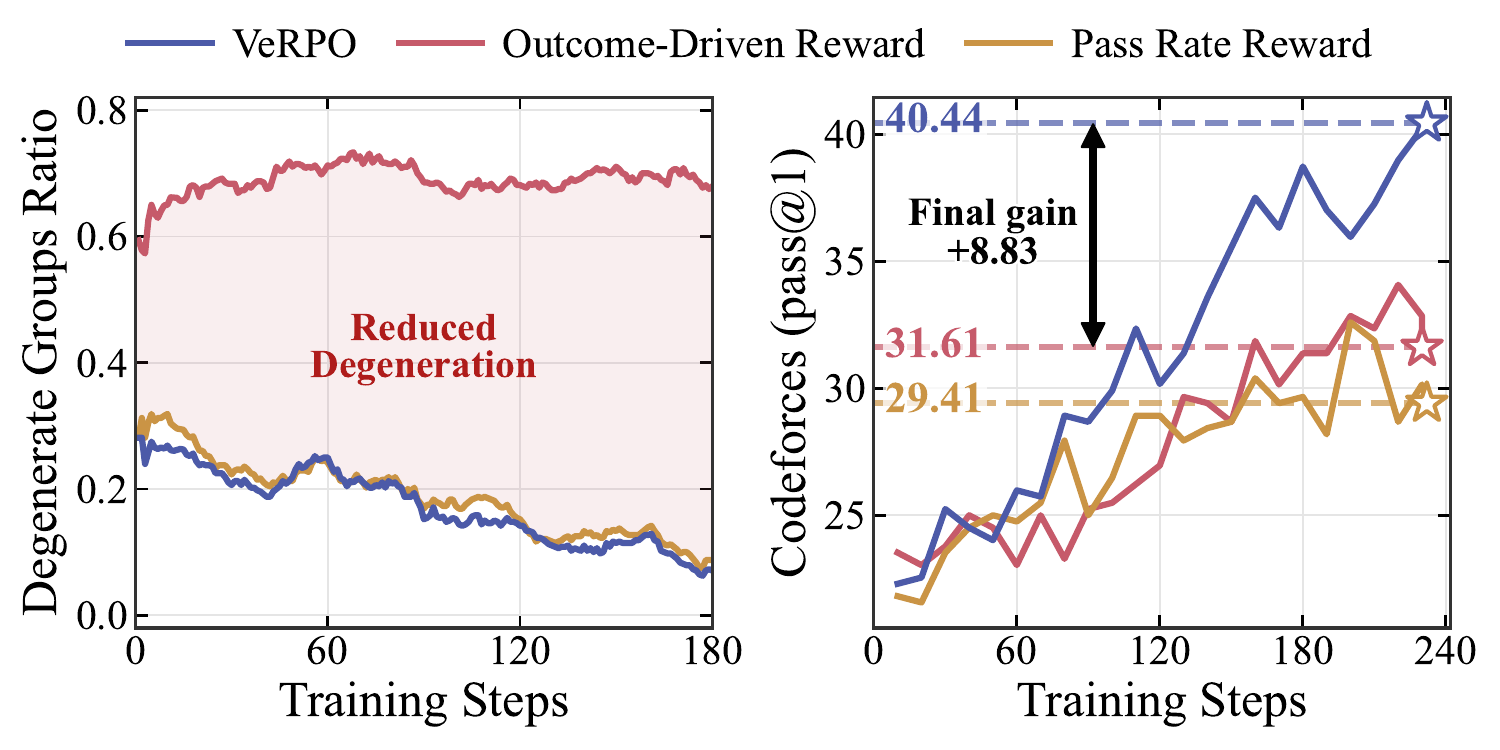}
  \caption{Evaluation of different verifiable reward designs. \textit{Left}: Signal efficiency, where the degenerate group ratio indicates the fraction of zero-advantage groups per batch. \textit{Right}: Pass@1 performance of group-based RL on Codeforces problems using Qwen3-8B.}
  \label{fig:intro1}
\end{figure}

Current reward designs for code generation predominantly fall into two distinct paradigms, each facing specific limitations. First, mainstream outcome-driven approaches enforce functional correctness\footnote{Functional correctness indicates that the generated code produces the expected output for all valid inputs.} with a verifiable binary reward derived from test-suite execution outcomes (e.g., 1 if all unit tests pass and 0 otherwise). 
However, such strict verification induces severe sparsity: when all sampled solutions receive identical binary outcomes, group-based RL methods such as GRPO \cite{guo2025deepseek} yield zero relative advantages and thus no gradient signal.
Alternatively, recent studies have explored external Reward Models (RMs) to provide dense supervision based on code semantics \cite{zeng-etal-2025-acecoder,ye2025process,dai2024processreward}, but external learned RMs introduce prohibitive computational overhead and vulnerability to reward hacking \cite{zhang2025reward-hacking}. 
This dual-bottleneck scenario raises a central question: \textit{Can we obtain dense rewards for code-generation RL while preserving verifiability and avoiding auxiliary reward models?}

The test-suite structure of programming problems suggests a promising answer: \textit{partial success} (i.e., passing a subset of test cases). Although outcome-driven rewards collapse execution feedback into a single binary signal, each generated solution actually receives multiple test-case-level outcomes\footnote{Code generation benchmarks typically contain substantial test cases, making partial success naturally dense; see Appendix~\ref{code_dataset_statistics} for dataset-level statistics.}. Passing a subset of tests is thus not equivalent to complete failure; rather, it provides verifiable evidence of intermediate functional correctness and offers denser and more informative signal without any auxiliary reward model. Fig.~\ref{fig:intro1}~(\textit{Left}) shows that even a naive partial-success reward---the pass rate of the test suite---substantially reduces the fraction of zero-advantage groups over outcome-driven rewards, indicating that partial success effectively densifies supervision.

However, a denser reward is not necessarily better. Although the naive pass-rate reward is both dense and verifiable, it can still degrade final performance compared to binary outcome-driven rewards (Fig.~\ref{fig:intro1}~\textit{Right}). This contrast suggests that partial success does not automatically yield an effective reward signal; it must be carefully calibrated.

In this paper,
we propose VeRPO, an RL framework that systematically turns verifiable partial success into dense and robust rewards. 
Rather than proposing another heuristic partial-success reward, we first introduce a general weighted-sum formulation over per-test execution results, which subsumes a broad class of partial-success rewards and provides an analytical lens for examining their optimization behavior.
Using this formulation, we reveal a severe \textbf{cardinality bias} inherent in rewarding partial success: redundant, easy tests dominate the optimization baseline by their sheer count, causing policy optimization to disproportionately favor gains from easy-test successes rather than progress on frontier tests. Building on this analysis, VeRPO introduces a dynamic, density-calibrated local reward that corrects the cardinality bias and provides informative dense supervision from partial success. To anchor optimization to end-to-end functional correctness, VeRPO further incorporates a global outcome reward derived from complete test-suite execution. Together, the two reward signals enable dense and reliable optimization without auxiliary reward models, yielding strong performance gains of VeRPO (Fig.~\ref{fig:intro1}). 
In diverse code generation benchmarks and settings, extensive experiments demonstrate that VeRPO consistently outperforms outcome-driven and RM-based baselines, achieving up to 8.83 performance gain in pass@1 with negligible time cost (< 0.02\%) and zero GPU memory overhead. 

Our key contributions are as follows:
\begin{itemize}
\item We systematically investigate the mechanism of rewarding partial success, demonstrating its potential for dense supervision while revealing an inherent cardinality bias that misguides policy optimization.
\item We propose VeRPO, an RL framework that corrects the cardinality bias in partial-success rewards and enables dense and reliable optimization purely from verifiable execution feedback, without auxiliary reward models.
\item Extensive experiments on diverse code-generation benchmarks and settings demonstrate the superior performance of VeRPO over baselines, with zero GPU memory overhead and negligible time cost.
\end{itemize}

\section{Related Work}

\subsection{RL for Code Generation}
RL has emerged as a central paradigm for improving LLM-based code generation \cite{liu2023rltf,wang2024enhancing-rlwithcode, zhang2025surveyreinforcementlearninglarge-rlwithcode}. Early methods, including CodeRL \cite{le2022coderl} and PPOCoder \cite{shojaee2023execution-ppocoder}, employ execution feedback as reward signals under the REINFORCE algorithm \cite{williams1992Reinforce-algo} and Proximal Policy Optimization framework (PPO; \citealp{schulman2017ppo}), respectively, establishing an early foundation for RL in this domain. More recent advances, exemplified by DeepSeek-R1 \cite{guo2025deepseek}, combine scalable RL with chain-of-thought reasoning, boosting interest in RL for code generation. Recent work has further extended RL to multi-turn code generation, where methods such as $\mu\text{CODE}$ \cite{jain2025multiturnthrougOne} and RLEF \cite{gehring2024rlef} support iterative self-correction for challenging coding problems. 
In line with the above studies, we focus on competitive code-generation tasks in this paper, where each model response is a complete solution candidate for a standalone programming problem and is evaluated against the same task-level test suite. This setting differs from agentic coding, which often presupposes a code-capable model and focuses on higher-level workflows such as planning, tool use, and environment interaction for software-engineering tasks \cite{jimenez2024swe,yang2024swe}.

\subsection{Reward Design in Code Generation}
Reward design plays a central role in RL for code generation \cite{sun2025large-rewarddesign,xu2025policy-rewarddesign}. 
Mainstream approaches typically rely on outcome-driven rewards computed from test-suite execution results.
For example, GRPO and its variants \cite{yu2025dapo, yue2025vapo, ahmadian2024back-leave-one-out} use binary execution outcomes as verifiable reward signals, directly aligning policy outputs with functional correctness.
More recently, external reward models (RMs) have been introduced to provide richer and denser supervision. For instance, AceCoder \cite{zeng-etal-2025-acecoder} trains a code-specific RM on large-scale preference pairs for fine-grained quality assessment, while CodePRM \cite{li2025codeprm} trains an RM on execution traces to provide dense supervision for intermediate reasoning steps. However, both paradigms face inherent limitations: outcome-driven rewards enforce functional correctness but introduce severe sparsity, while RM-based methods offer dense supervision but suffer from misalignment and computational overhead.

Different from these lines of work, we focus on the verifiable partial-success structure present in code execution itself. Rather than relying only on binary terminal outcomes or introducing auxiliary learned reward models, we systematically study how partial success can be translated into reliable reward signals for code-generation RL.

\section{Preliminaries}
\label{sec:preliminaries}
\subsection{Code Generation as an MDP}\label{MDP}
We formulate code generation as a Markov Decision Process (MDP),  where a policy $\pi_\theta$ generates up to $T$ code candidates and receives execution feedback after each attempt. Given a problem prompt $x \in \mathcal{X}$, at each turn $t \in \{1, \dots, T\}$, the policy generates a code candidate $y_t$ conditioned on the current state $s_t = (x, y_1, o_1, \ldots, y_{t-1}, o_{t-1})$, which records the interaction history up to turn $t$, with $s_1 = x$.
After generating $y_t$, the code is executed against the test suite $\mathcal{U}_x = \{u_1, \dots, u_{|\mathcal{U}_x|}\}$ and yields the corresponding execution feedback $o_t$. The process terminates once the generated code passes the test suite $\mathcal{U}_x$ or the maximum turn budget $T$ is reached, yielding a trajectory $\tau = \big((s_1, y_1, o_1), (s_2, y_2, o_2), \dots, (s_{|\tau|}, y_{|\tau|}, o_{|\tau|})\big)$, where $|\tau|$ denotes the number of executed turns. This formulation naturally unifies both single-turn ($T=1$) and multi-turn ($T>1$) code generation.

\subsection{Group-Based RL}
Group-based RL has become a prominent paradigm for optimizing LLMs. Given a prompt $x$, a group of $N$ trajectories $\mathcal{G}_x = \{\tau_1, \tau_2, \ldots, \tau_N\}$ is sampled under $\pi_{\theta_{\text{old}}}$. Each trajectory $\tau_i$ receives a scalar reward $R(\tau_i)$ indicating its quality (e.g., 1 for correct, 0 for incorrect). 
Unlike standard actor-critic frameworks such as PPO, which require a separate value network to compute optimization baseline for advantage estimation, group-based RL methods estimate advantages directly from the rewards within the sampled group. Typically, they utilize the group mean reward as the optimization baseline, defining the advantage for each trajectory as:
\begin{equation}
A(\tau_i) = \left( R(\tau_i) - \operatorname{mean}\left( \{R(\tau_j)\}_{j=1}^N \right) \right) \;/\; F_{\text{norm}}.
\label{eq:grpo}
\end{equation}
where $F_{\operatorname{norm}}$ denotes the advantage normalization factor. GRPO defaults to $F_{\text{norm}} = \operatorname{std}$, whereas such normalization introduces task-level difficulty bias \cite{liu2025understandingr1-zero-likeTraining}.
Unless otherwise specified, we set a constant normalization factor $F_{\text{norm}} = 1$, which yields an unbiased Leave-One-Out estimator \cite{feng2025groupingroup,kool2019buy-leave-one-out}.

\section{Method}\label{method}
This section presents the analysis that motivates VeRPO and the reward design that implements it. 
Section~\ref{Partial Success and Cardinality Bias} formalizes partial-success rewards and reveals an inherent cardinality bias that misguides policy optimization. 
Section~\ref{VeRPO} then introduces VeRPO, which corrects this bias and turns partial success into dense, reliable rewards.

\subsection{Partial Success and Cardinality Bias}\label{Partial Success and Cardinality Bias}
In code generation tasks with multiple test cases, each generated solution is verified against all test cases, yielding a vector of per-test pass/fail outcomes at each turn. The central question is how to turn this partial-success information into an effective reward for RL training. Our formulation is guided by a dual consideration of generality and simplicity: for generality, test cases differ in hardness, thus their varying contributions to the reward need to be explicitly modeled; for simplicity, we do not model cross-turn credit assignment—since each turn produces a complete code solution evaluated against the full test suite, the execution feedback at each turn directly measures the functional quality of that solution and requires no temporal attribution across turns.
\textcolor{black}{Consequently, we quantify this turn-level partial success as a weighted sum over per-test outcomes,  yielding a non-binary reward signal across both single- and multi-turn settings:}
\begin{equation}
    r_{t,i} = \sum\nolimits_{j=1}^{|\mathcal{U}_x|} w_j\, p^{(j)}_{t,i},\label{eq:reward'}
\end{equation}
where $p^{(j)}_{t,i} \in \{0,1\}$ denotes whether the turn-$t$ code in trajectory $\tau_i$ passes test case $u_j \in \mathcal{U}_x$, and $w_j\geq 0$ specifies the contribution of test case $u_j$. 
\textcolor{black}{
Eq.~\eqref{eq:reward'} allows flexible modeling of test-level contributions, with the standard pass-rate reward serving as the uniform-weight special case.}

\begin{figure}[t]
  \includegraphics[width=\columnwidth]{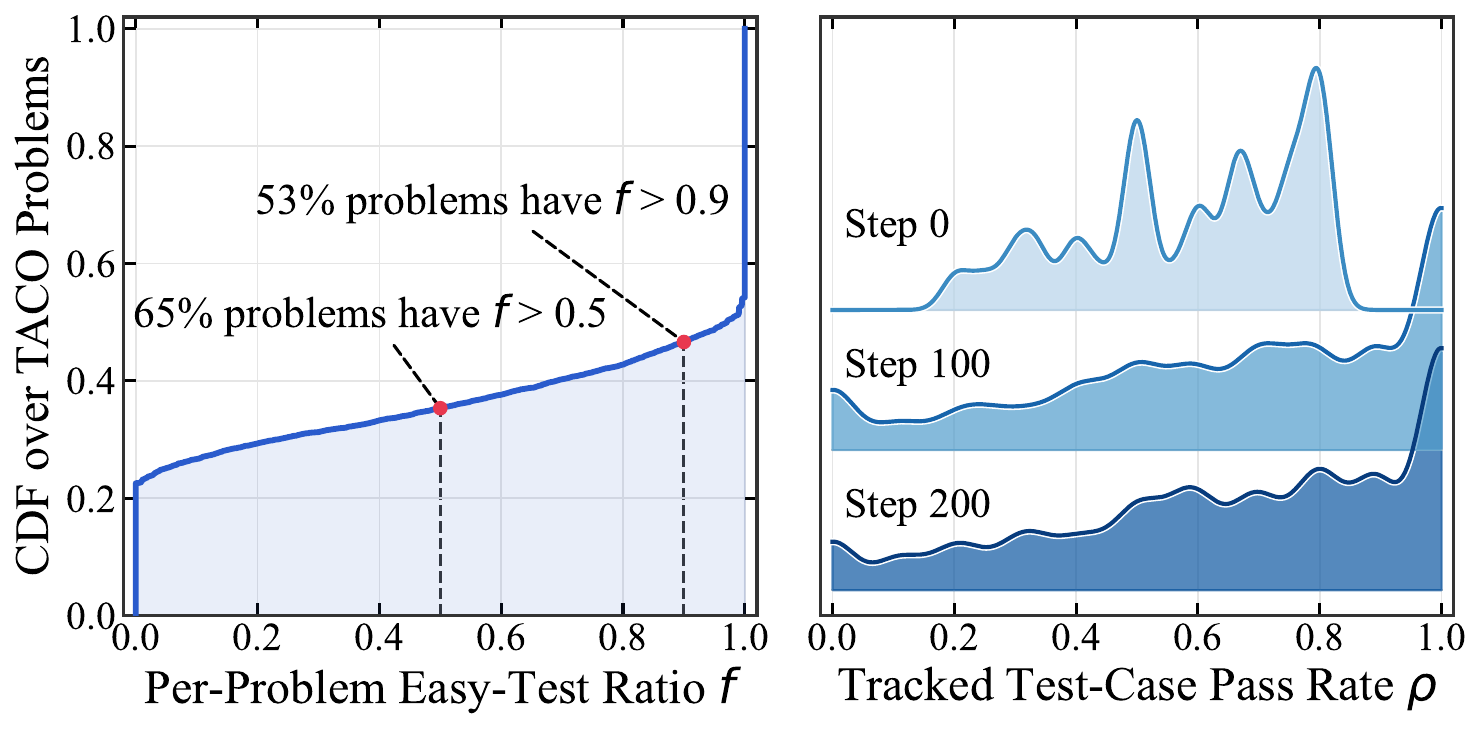}
  \caption{
  Test-case pass-rate skew on TACO with Qwen3-8B. \textit{Left}: CDF over TACO problems of $f$, where $f$ denotes the fraction of test cases in $\mathcal{U}_x$ with pass rate $\rho > 0.8$. 
  \textit{Right}: Pass-rate distribution during training for test cases with $\rho \in [0.2, 0.8]$ at Step 0.
  }
  \label{fig:method1}
\end{figure}

\paragraph{Cardinality bias in the optimization baseline.}

One key aspect of implementing Eq.~\eqref{eq:reward'} is how to assign weights properly to each test case. Rather than simply relying on heuristics, we identify a structural skewness in the difficulty distribution of test cases and formally show how it induces an intrinsic bias in the optimization baseline—which we term \textbf{cardinality bias}. This in turn gives us a principled foundation for weight design.

\begin{figure*}[t]
  \centering
  \includegraphics[width=0.9\textwidth]{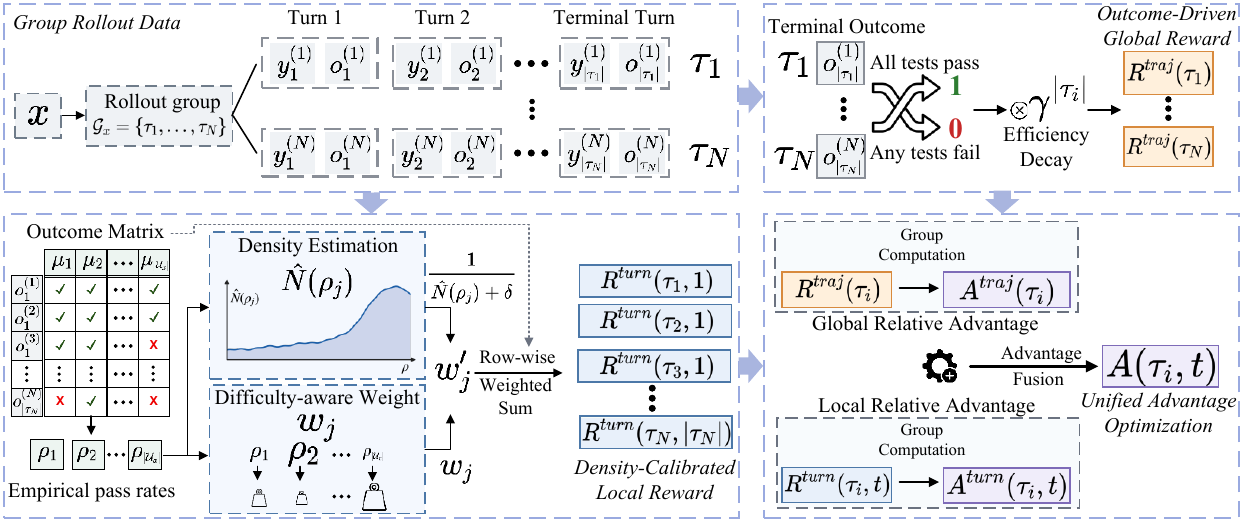}
  \caption{Overview of VeRPO. VeRPO fuses density-calibrated local rewards derived from partial success with outcome-driven global rewards, enabling effective and dense policy optimization for code generation.}
  \label{fig:method}
\end{figure*}

The structural skewness arises from the uneven difficulty distribution of test cases. We measure test-case difficulty by empirical pass rate---the fraction of generated code passing a given test. Formally, for problem $x$ with test suite $\mathcal{U}_x$, the empirical pass rate of test case $u_j \in \mathcal{U}_x$ over rollout group $\mathcal{G}_x$ is:
\begin{align}
\rho_j = \frac{1}{M_x}\sum_{i=1}^{|\mathcal{G}_x|}\sum_{t=1}^{|\tau_i|} p^{(j)}_{t,i}, \quad M_x = \sum_{i=1}^{|\mathcal{G}_x|}|\tau_i|,
\end{align}
where $M_x$ is the total number of turns in trajectory group $\mathcal{G}_x$. A test case is easy if $\rho_j$ is close to 1, and in practice, easy test cases constitute the majority. 
\textcolor{black}{As shown in Fig.~\ref{fig:method1}~(\textit{Left}), the difficulty distribution of test cases is highly skewed before training begins: for 53\% of TACO problems with Qwen3-8B, over 90\% of their respective test cases are already easy ($\rho_j>0.8$). Moreover, as RL training proceeds, this skewness is further exacerbated, with previously intermediate-difficulty tests ($\rho_j\in[0.2,0.8]$) progressively migrating into the high-pass-rate region (Fig.~\ref{fig:method1},~\textit{Right}). Similar results on other datasets are shown in Appendix~\ref{app:additional_skew}.}

\textcolor{black}{To formally analyze how the uneven difficulty distribution induces a bias, \textcolor{black}{we examine the optimization baseline in group-based RL, which corresponds to the group mean in Eq.~\eqref{eq:grpo}. Since Eq.~\eqref{eq:reward'} defines a turn-level reward signal, we instantiate its group-mean baseline at the same granularity:}
\begin{align}
\bar r^{\text{turn}}
&=
\frac{1}{M_x}
\sum\nolimits_{i=1}^{|\mathcal G_x|}
\sum\nolimits_{t=1}^{|\tau_i|}
r_{t,i}.\label{eq.basline_1}
\end{align}
Next, substituting Eq.~\eqref{eq:reward'} into Eq.~\eqref{eq.basline_1} gives
\begin{align}
    \bar r^{\text{turn}}
&=
\frac{1}{M_x}
\sum\nolimits_{i=1}^{|\mathcal G_x|}
\sum\nolimits_{t=1}^{|\tau_i|}
\sum\nolimits_{j=1}^{|\mathcal U_x|}
w_j p_{t,i}^{(j)} \notag\\
&=
\sum\nolimits_{j=1}^{|\mathcal U_x|}
w_j
\underbrace{\left(
\frac{1}{M_x}
\sum\nolimits_{i=1}^{|\mathcal G_x|}
\sum\nolimits_{t=1}^{|\tau_i|}
p_{t,i}^{(j)}
\right)}_{ \text{empirical pass rate of }j\text{-th test}} \notag\\
&=
\sum\nolimits_{j=1}^{|\mathcal U_x|} w_j \rho_j.
    \label{eq:baseline_2}
\end{align}
Eq.~\eqref{eq:baseline_2} reveals that the optimization baseline equals the weighted sum of per-test empirical pass rates, which can be rewritten in integral form (derivation in Appendix~\ref{Derivation_density_form}):
\begin{equation}
    \bar{r}^{\,\text{turn}}=\int_0^1 \bar{w}(\rho)\,\rho\,N(\rho)\,d\rho.
    \label{eq:density_form}
\end{equation}
where $\bar{w}(\rho)$ is the average weight over tests with its pass rate equal to $\rho$, and
$N(\rho)=\sum_{j=1}^{|\mathcal{U}_x|}\delta(\rho-\rho_j)$ is the empirical density of test cases at pass rate $\rho$---precisely the difficulty distribution.}

Recall that $N(\rho)$ is heavily skewed toward the high-pass-rate end (Fig.~\ref{fig:method1}). The high-$\rho$ region therefore dominates the baseline contribution in Eq.~\eqref{eq:density_form} by sheer count, making easy tests disproportionately influential despite the designed weights $\{w_j\}$ (see Appendix~\ref{app:contribution_gap_bound} for formal analysis). As a result, a partial-success reward $r_{t,i}$ tends to exceed the inflated baseline primarily by passing more easy tests than average---progress on hard frontier tests contributes far less toward pushing the reward above the baseline. Policy updates therefore preferentially reinforce easy-test gains rather than progress at the model's capability boundary. We term this \textbf{cardinality bias}.

\subsection{VeRPO}\label{VeRPO}
We propose VeRPO, a group-based RL framework that addresses the cardinality bias analyzed in Section~\ref{Partial Success and Cardinality Bias} and retains dense, robust supervision from partial success.
VeRPO comprises two complementary components: a \emph{density-calibrated local reward} that corrects cardinality bias to extract informative dense supervision from partial success, and an \emph{outcome-driven global reward} that anchors optimization to holistic functional correctness. The overview of VeRPO is depicted in Fig.~\ref{fig:method}.

\paragraph{Density-Calibrated Local Reward.}
Cardinality bias arises because $N(\rho)$ in Eq.~\eqref{eq:density_form} amplifies the baseline contribution of dense easy-test regions regardless of the designed weights. VeRPO corrects this by estimating $N(\rho)$ at each $\rho_j$ via a Gaussian kernel density estimator:
\begin{equation}
\hat{N}(\rho_j) = \sum\nolimits_{j'=1}^{|\mathcal{U}_x|} \exp\!\left(-(\rho_j - \rho_{j'})^2 \big/ 2\sigma^2\right),
\end{equation}
where $\sigma$ is the kernel bandwidth.
Dividing each test-case weight by $\hat{N}(\rho_j)$ counteracts the local density factor $N(\rho)$ in Eq.~\eqref{eq:density_form}, yielding an approximation to the density-free baseline $\int_0^1 \bar{w}(\rho)\,\rho\,d\rho$ and ensuring it is governed by the designed weights rather than by the cardinality of easy tests. Beyond density correction, since $\rho_j$ naturally reflects test difficulty under the current policy, VeRPO further assigns an exponential difficulty-aware weight:
\begin{equation}
w_j = \exp(-\alpha \rho_j),\label{eq:w_J}
\end{equation}
where $\alpha > 0$ controls the degree of emphasis on rarely-passed frontier tests relative to already-mastered ones. Combining density calibration and difficulty-aware weighting, the final adjusted weight for test case $u_j$ is:
\begin{equation}
{w}_j' = {\exp(-\alpha \rho_j)} \big/ ({\hat{N}(\rho_j) + \delta}).
\end{equation}
where $\delta > 0$ is a small constant for numerical stability. The local partial-success reward for turn $t$ in trajectory $\tau_i$ is thus:
\begin{equation}
R^{turn}(\tau_i, t) = \sum\nolimits_{j=1}^{\small{|\mathcal{U}_x|}} w_j' \cdot p_{t,i}^{(j)},\label{reward}
\end{equation}
which focuses optimization on the unmastered frontier rather than the redundant majority.

\paragraph{Outcome-Driven Global Reward.}
While the density-calibrated local reward provides dense supervision, it optimizes over partial execution outcomes and does not directly target whether the complete test suite passes. To anchor optimization to end-to-end functional correctness, VeRPO further incorporates a binary trajectory-level outcome reward
\(
R^{traj} \in \{0,1\},
\)
where $R^{traj}(\tau_{i}) = 1$ if the terminal execution feedback passes the complete test suite, and $R^{traj}(\tau_{i}) = 0$ otherwise. Since this reward is defined on the holistic test-suite outcome rather than an aggregation of per-test partial success, it is free from the cardinality bias analyzed in Section~\ref{Partial Success and Cardinality Bias}.
We further apply an efficiency-aware decay to favor trajectories that reach correct solutions in fewer turns:
\begin{equation}
\tilde{R}^{traj}(\tau_{i}) = R^{traj}(\tau_{i}) \cdot \gamma^{|\tau_{i}|},
\end{equation} 
where \(\gamma\in(0,1]\) controls the strength of the efficiency preference. Together, the local and global rewards provide complementary supervision: the former densifies the training signal through calibrated partial success, while the latter preserves strict alignment with functional correctness.

\newcommand{\dashfill}{\leavevmode\cleaders\hbox to 0.5em{\hss-\hss}\hfill\kern0pt}

\begin{table*}[t]
\centering
\fontsize{8.8pt}{9.68pt}\selectfont
\setlength{\tabcolsep}{2.2pt} 
\begin{tabular}{*{18}{c}}
\toprule
\multirow{3}{*}{Method} & \multirow{3}{*}{Train} & \multirow{3}{*}{RM} & \multirow{3}{*}{Reward} & \multicolumn{4}{c}{HumanEval} & \multicolumn{4}{c}{BigCodeBench} & \multicolumn{2}{c}{LCB} & \multicolumn{2}{c}{Codeforces} & \multicolumn{2}{c}{\multirow{2}{*}{Avg}} \\ 
\cmidrule(lr){5-8} \cmidrule(lr){9-12} \cmidrule(lr){13-14} \cmidrule(lr){15-16}
& & & & \multicolumn{2}{c}{-} & \multicolumn{2}{c}{Plus} & \multicolumn{2}{c}{Full} & \multicolumn{2}{c}{Hard} & \multicolumn{2}{c}{V6} & \multicolumn{2}{c}{CodeElo} & \multicolumn{2}{c}{} \\ 
\cmidrule(lr){5-6} \cmidrule(lr){7-8} \cmidrule(lr){9-10} \cmidrule(lr){11-12} \cmidrule(lr){13-14} \cmidrule(lr){15-16} \cmidrule(lr){17-18}
& & & & ST & MT & ST & MT & ST & MT & ST & MT & ST & MT & ST & MT & ST & MT \\ 
\midrule

Qwen3-8B & - & - & - & 91.03 & 95.05 & 87.49 & 90.14 & 35.85 & 58.32 & 16.38 & 41.69 & 27.12 & 28.14& 20.77 & 22.73 & 46.44 & 56.01 \\ 
\midrule

\multirow{2}{*}{GRPO} & ST & - & 0-1 & {91.99} & 96.34 & 88.03& 90.32 & \uline{36.33} & 57.87 & 16.87 & 40.54 & 28.35 & 29.28 & 28.49 & 30.72 & 48.34 & 57.51 \\
                      & MT & - & 0-1 & 91.46 & 96.41 &87.57 & 91.23 & 35.73 & \uline{60.91} & 16.55 & \uline{43.32} & 28.35 & \uline{30.50} & 27.38 & 31.61 & 47.84& \uline{59.00} \\
\addlinespace 

\multirow{2}{*}{AceCoder} & ST & 7B & Dense & \uline{92.04} & \uline{96.59} & \uline{88.71}& 91.46 & 36.07 & 58.36 & 14.86 & 39.53 & 27.78 & 27.57 &25.06 &26.96 & 47.42 & 56.74 \\
                          & MT & 7B & Dense & \multicolumn{14}{c}{\dashfill Training Collapse \dashfill} \\
\addlinespace

\multirow{2}{*}{VeRPO} & ST & - & Dense & \textbf{92.73} & {96.49} & \textbf{89.32}& \uline{91.84} & \textbf{37.42}& 58.95 & \textbf{17.91} & 41.55& \textbf{29.72} & 30.14 & \textbf{30.50} & \uline{33.76} & \textbf{49.60} & 58.79 \\
                        & MT & - & Dense & 91.69 & \textbf{97.87} & {88.05} & \textbf{93.14} & 36.10 & \textbf{62.52} & \uline{17.22} & \textbf{45.69} & \uline{29.07} & \textbf{33.07} & \uline{29.03} & \textbf{40.44} & \uline{48.53} & \textbf{62.12} \\

\bottomrule
\end{tabular}
\caption{Performance comparison on six benchmarks using pass@1. ST/MT denotes single-turn/multi-turn setting for training and evaluation. Dashed entries denote training collapse due to optimization instability. Best/second-best results for each benchmark and evaluation setting are highlighted in \textbf{bold}/\uline{underlined}.}
\label{tab:my_results_restructured}
\end{table*}
\paragraph{Unified Advantage Optimization.}
To combine the two reward signals in a unified policy update, VeRPO constructs a local and a global advantage, and fuses them into a turn-level training objective.
For the local reward, we collect all turn-level partial-success rewards within the rollout group as
\(  
G(\mathcal{G}_x)
= \{R^{turn}(\tau_i,t) \mid \tau_i\in\mathcal{G}_x,\ 1\le t\le |\tau_i|\},
\)
and define the local relative advantage as
\begin{equation}
  A^{turn}(\tau_i, t) = {R^{turn}(\tau_i, t) - \operatorname{mean}(G(\mathcal{G}_x))}.
\end{equation}
At the trajectory level, we compute a global relative advantage from the decayed outcome reward: 
\begin{equation}
A^{traj}(\tau_{i}) = {\tilde{R}^{traj}(\tau_{i}) - \operatorname{mean}(\{\tilde{R}^{traj}(\tau_{j})\}_{j=1}^N)}.
\end{equation}
We then broadcast the trajectory-level advantage to each turn and combine it with the local advantage:
\begin{equation}
A(\tau_i, t) = A^{traj}(\tau_i) + \beta \cdot A^{turn}(\tau_i, t),
\end{equation}
where \(\beta \ge 0\) controls the contribution of the two advantages. The trajectory-level advantage $A^{traj}(\tau_i)$ anchors the global optimization, ensuring policy updates adhere to end-to-end functional correctness, while the turn-level advantage $A^{turn}(\tau_i, t)$ enriches the learning signal with dense feedback from local partial success.
Finally, VeRPO optimizes the policy with a clipped group-based objective using the unified turn-level advantage $A(\tau_i,t)$; the full objective is provided in Appendix~\ref{app:verpo_objective}.

\section{Experiments}

\subsection{Setup}

\paragraph{Datasets.}  For training, we utilize the subset dataset derived from \cite{luo2025deepcoder}, comprising 7.4K verified code problems from TACO \cite{li2023taco}. We evaluate performance across four mainstream code generation benchmark suites: 1) HumanEval \cite{chen2021humaneval} along with its enhanced variant HumanEval-Plus \cite{liu2023humanevalplus}; 2) BigCodeBench \cite{zhuo2024bigcodebench} reporting results on both its Full and Hard subsets; 3)  LiveCodeBench (LCB, V6, 2023.05--2025-04) \cite{jain2024livecodebench}; and 4) Codeforces problems derived from CodeElo \cite{quan2025codeelo}. 

\paragraph{Baselines.}  We benchmark VeRPO against the predominant group-based method, GRPO, instantiated with two distinct reward configurations: 1) \textit{vanilla outcome-driven rewards}, representing the standard GRPO configuration that relies on binary execution outcomes indicating overall functional correctness; and 2) \textit{RM-based dense rewards},  which leverages an external RM to assign fine-grained rewards to intermediate turns. Following prior work, we employ AceCodeRM-7B \cite{zeng-etal-2025-acecoder} to generate turn-level dense rewards.

\paragraph{Implementation.}  We use Qwen3-8B as the main RL backbone, with additional backbone-scale results reported in Appendix~\ref{sec:appendix_qwen3_4b_main_results}. To evaluate performance across varying horizons, we conduct training and evaluation in both single-turn (ST, $T=1$) and multi-turn (MT, $T=4$) code generation settings. During training, we utilize a rollout batch size of 32 code problems with 10 responses sampled per problem; maximum response length is set to 16,384 tokens and sampled with temperature 1.0. For evaluation, we set the maximum response length to 16,384 tokens and sampling temperature to 0.6. We report pass@1, computed with the unbiased estimator of \cite{chen2021humaneval}, as the primary functional-correctness metric. See Appendix~\ref{sec:appendix_training_evaluation_details} for additional training and evaluation details.

\subsection{Results and Analysis}
\subsubsection{Performance of VeRPO}\label{Performance}
Table~\ref{tab:my_results_restructured} reports pass@1 results across six benchmarks under single-turn (ST) and multi-turn (MT) training/evaluation settings.
In the ST$\rightarrow$ST setting, VeRPO achieves the best performance across all benchmarks, obtaining an average improvement of +1.26 and +2.18 over the outcome-driven GRPO and the RM-based AceCoder, respectively. The superiority underscores the intrinsic efficacy of VeRPO, yielding stronger effectiveness in the standard single-response setting.
The performance gain of VeRPO becomes more pronounced in the MT setting. AceCoder fails to complete stable MT training due to optimization collapse, as detailed in Appendix~\ref{sec:appendix_training_stability_analysis}, suggesting the instability risk of relying on an external black-box reward model for iterative optimization. In contrast, VeRPO is fully grounded in verifiable execution feedback and remains stable during optimization. Compared with execution-based GRPO under MT training and evaluation, VeRPO improves the average pass@1 by 3.12 points, with the largest gain on the challenging Codeforces benchmark (+8.83). This margin indicates that extracting effective dense learning signals from partial success contributes to improved performance on harder problems. Furthermore, VeRPO retains strong cross-setting performance, outperforming the corresponding baselines even under mismatched training and evaluation horizons. 

\begin{table}[t]
\centering
\fontsize{9.5pt}{9.9pt}\selectfont
\setlength{\tabcolsep}{1.5pt}
\begin{tabular}{cccccc} 
\toprule
\multirow{2}{*}{Method} & HumanEval & \multicolumn{2}{c}{BigCodeBench} & LCB & Codeforces \\ 
\cmidrule(lr){2-2} \cmidrule(lr){3-4} \cmidrule(lr){5-5} \cmidrule(lr){6-6}
 & Plus & Full & Hard & V6 & CodeElo \\ 
\midrule
w/o $A^{turn}$ & 91.58 & 61.58 & 44.25 & 30.64 & 34.68 \\
w/o $A^{traj}$ & 92.31 & 62.12 & 45.18 & 31.14 & 37.62 \\
w/ std & 93.06 & 61.87 & 44.67 & 31.64 & 38.05 \\
\midrule
VeRPO & \textbf{93.14} & \textbf{62.52} & \textbf{45.69} & \textbf{33.07} & \textbf{40.44} \\ 
\bottomrule
\end{tabular}
\caption{Ablation study on components of VeRPO.}
\label{tab:Ablation-Study}
\end{table}

\begin{table}[t]
\centering
\fontsize{8.8pt}{9.79pt}\selectfont
\setlength{\tabcolsep}{1.15pt}
\begin{tabular}{ccccccc}
\toprule
\multirow{2}{*}{$A^{traj}$} & \multirow{2}{*}{$R^{turn}$} & HumanEval & \multicolumn{2}{c}{BigCodeBench} & LCB & Codeforces \\ \cmidrule(lr){3-3} \cmidrule(lr){4-5} \cmidrule(lr){6-6} \cmidrule(lr){7-7}
 &  & Plus & Full & Hard & V6 & CodeElo \\ \midrule
 GRPO & - & 91.23 & 60.91 & 43.32 & 30.50 & 31.61\\ \midrule

- & PS & 91.69 & 59.45 & 43.49 & 29.64 & 29.41 \\
- & Diff & 91.46 & 59.78 & 41.55 & 30.28 & 35.23 \\
- & VeRPO & 92.31 & 62.12 & 45.18 & 31.14 & 37.62 \\\midrule
  VeRPO & PS & 92.14 & 61.23 & 44.34 & 32.00 & 35.75 \\
  VeRPO & Diff & 91.31 & 60.00 & 43.15 & 30.57 & 35.11 \\ 
  VeRPO & VeRPO & \textbf{93.14} & \textbf{62.52} & \textbf{45.69} & \textbf{33.07} & \textbf{40.44} \\ \bottomrule
  \end{tabular}
  \caption{Ablation study on local partial-success reward designs for constructing the turn-level reward $R^{turn}$. PS: raw pass-rate rewards; Diff: difficulty-weighted rewards without density normalization.}
  \label{tab:ab2}
  \end{table}
  
\subsubsection{Ablation Study}
\paragraph{Effect of Components in VeRPO.}  We ablate three components of VeRPO in the multi-turn setting: the local turn-level advantage $A^{turn}$, the global trajectory-level advantage $A^{traj}$, and the fixed normalization factor $F_{\operatorname{norm}}=1$. We compare the full model with variants that remove $A^{turn}$, remove $A^{traj}$, or replace fixed normalization with standard-deviation normalization.
Table~\ref{tab:Ablation-Study} shows that removing any component degrades performance. Removing $A^{turn}$ leads to the largest degradation, with consistent drops across all benchmarks and a particularly large decline on Codeforces, from 40.44 to 34.68, which is consistent with the role of $A^{turn}$ in providing dense partial-success supervision beyond sparse outcome feedback. Removing $A^{traj}$ also lowers performance across all benchmarks, indicating that trajectory-level correctness remains important for anchoring optimization to full functional correctness. Finally, replacing the fixed normalization with standard-deviation normalization consistently underperforms VeRPO, but the gap is smaller than those caused by removing either advantage term. This suggests that fixed normalization improves calibration, while the dominant gains come from combining local turn-level and global trajectory-level advantages.

\paragraph{Ablation on Partial-Success Reward Designs.}
To rigorously validate the design of the local partial-success reward $R^{turn}$ in VeRPO, we benchmark VeRPO against two execution-based partial-success rewards:  (a) \textit{Raw Pass Rate} (PS), which uses the fraction of passed unit tests and ignores test difficulty; and (b) \textit{Difficulty-Weighted} (Diff), which applies the difficulty-aware weight in Eq.~\ref{eq:w_J} but removes the kernel density normalization. We evaluate each reward both in isolation and when fused with the global trajectory-level advantage $A^{traj}$.
Table~\ref{tab:ab2} shows that uncalibrated partial-success rewards are not sufficient. When used alone, both PS and Diff degrade performance relative to VeRPO and even underperform the outcome-driven GRPO baseline without turn-level partial-success rewards, especially on harder benchmarks such as BigCodeBench-Hard, LCB, and Codeforces. This indicates that dense partial-success feedback can be harmful if it is not properly calibrated.  Adding the global trajectory-level advantage improves these variants, but they still lag behind VeRPO. The remaining gap suggests that difficulty weighting alone does not address the cardinality bias identified in Section~\ref{Partial Success and Cardinality Bias}; density calibration is needed to make partial success a reliable learning signal. See Appendix~\ref{app:case_study} for rollout-level case studies.

\begin{figure}[t]
  \includegraphics[width=\columnwidth]{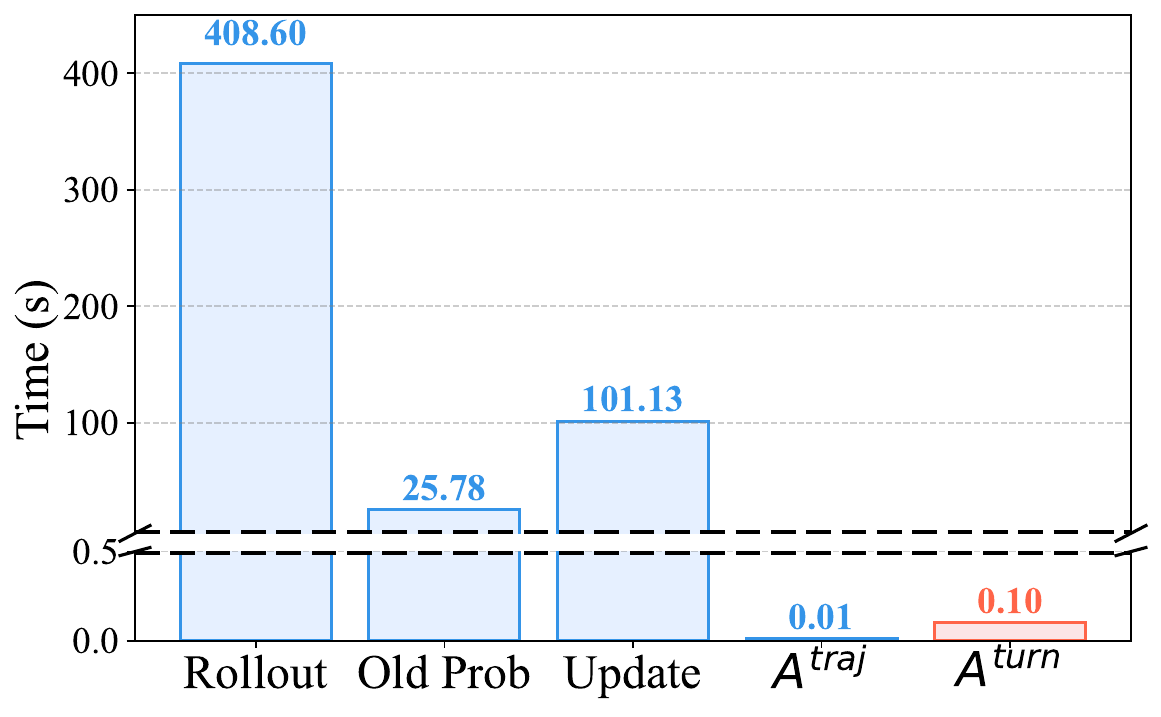}
  \caption{Computational analysis of VeRPO. Components shared with GRPO are highlighted in blue, while VeRPO-specific computations are shown in orange. A broken y-axis scale is employed to enhance the visualization of small values.}
  \label{fig:experiments2}
\end{figure}

\subsubsection{Signal Efficiency and Information Loss}
To better understand the performance gains of VeRPO, we analyze how often rollout groups provide non-degenerate optimization signals in the multi-turn setting. We employ the degenerate group ratio: the fraction of rollout groups with identical rewards, which yield zero relative advantages and thus no policy-gradient signal under group-relative optimization.
Fig.~\ref{fig:intro1} (\textit{Left}) shows that outcome-driven GRPO maintains a high degenerate group ratio, typically between 60\% and 70\%, indicating that binary outcome rewards discard substantial information from partial execution outcomes. In contrast, VeRPO keeps this ratio below 25\% for most of training and further reduces it as optimization proceeds, showing that partial success can be turned into more frequently usable training signals. Importantly, a lower degenerate group ratio alone does not guarantee better optimization: as discussed in Section~\ref{Introduction}, naive pass-rate rewards can densify supervision but still degrade final performance. The gains of VeRPO therefore come not merely from making the reward denser, but from calibrating partial-success signals through density-aware weighting while retaining the trajectory-level correctness anchor. 

\subsubsection{Computational Cost}
We assess the computational overhead introduced by VeRPO. As detailed in Section~\ref{method}, VeRPO follows the same group-based RL pipeline of GRPO, inheriting an identical grouped rollout sampling, computation of old log-probabilities, and clipped actor policy updates. Both pipelines eliminate the need for auxiliary critic or reward models by relying solely on execution feedback to compute rewards and advantages, thus sharing identical GPU memory consumption and LLM rollout overheads.

The only additional computation is the density-calibrated turn-level reward and advantage estimation.  Fig.~\ref{fig:experiments2} provides a per-iteration time breakdown. Computing $A^{turn}$ adds only 0.10s per iteration, accounting for less than 0.02\% of total training time, while rollout and policy updates dominate the cost. These results demonstrate that VeRPO achieves dense and effective supervision with virtually zero marginal computational cost, effectively defying the conventional trade-off between reward granularity and computational efficiency.

\section{Conclusion}
We introduce VeRPO, an RL framework that turns verifiable partial success into robust dense rewards for code generation. Our systematic analysis revealed an inherent cardinality bias in partial-success rewards: redundant easy tests dominate the optimization baseline by sheer count, causing policy updates to favor easy-test gains over frontier-test progress. Building on this analysis, VeRPO introduces a density-calibrated local reward that corrects the bias and combines it with a global outcome-driven reward that anchors optimization to end-to-end functional correctness. Extensive experiments across various benchmarks and settings demonstrate consistent gains over outcome-driven and RM-based baselines with negligible computational overhead.

\bibliography{custom}

@article{yang2025qwen3,
  title={Qwen3 technical report},
  author={Yang, An and Li, Anfeng and Yang, Baosong and Zhang, Beichen and Hui, Binyuan and Zheng, Bo and Yu, Bowen and Gao, Chang and Huang, Chengen and Lv, Chenxu and others},
  journal={arXiv preprint arXiv:2505.09388},
  year={2025}
}

@article{le2022coderl,
  title={Coderl: Mastering code generation through pretrained models and deep reinforcement learning},
  author={Le, Hung and Wang, Yue and Gotmare, Akhilesh Deepak and Savarese, Silvio and Hoi, Steven Chu Hong},
  journal={Advances in Neural Information Processing Systems},
  volume={35},
  pages={21314--21328},
  year={2022}
}

@article{guo2025deepseek,
  title={Deepseek-r1 incentivizes reasoning in llms through reinforcement learning},
  author={Guo, Daya and Yang, Dejian and Zhang, Haowei and Song, Junxiao and Wang, Peiyi and Zhu, Qihao and Xu, Runxin and Zhang, Ruoyu and Ma, Shirong and Bi, Xiao and others},
  journal={Nature},
  volume={645},
  number={8081},
  pages={633--638},
  year={2025},
  publisher={Nature Publishing Group UK London}
}

@article{luo2025deepcoder,
  title={Deepcoder: A fully open-source 14b coder at o3-mini level},
  author={Luo, Michael and Tan, Sijun and Huang, Roy and Patel, Ameen and Ariyak, Alpay and Wu, Qingyang and Shi, Xiaoxiang and Xin, Rachel and Cai, Colin and Weber, Maurice and others},
  journal={Notion Blog},
  year={2025}
}

@article{mroueh2025reinforcement-rlvrwithcode,
  title={Reinforcement Learning with Verifiable Rewards: GRPO's Effective Loss, Dynamics, and Success Amplification},
  author={Mroueh, Youssef},
  journal={arXiv preprint arXiv:2503.06639},
  year={2025}
}

@article{yue2025vapo,
  title={Vapo: Efficient and reliable reinforcement learning for advanced reasoning tasks},
  author={Yue, Yu and Yuan, Yufeng and Yu, Qiying and Zuo, Xiaochen and Zhu, Ruofei and Xu, Wenyuan and Chen, Jiaze and Wang, Chengyi and Fan, TianTian and Du, Zhengyin and others},
  journal={arXiv preprint arXiv:2504.05118},
  year={2025}
}

@article{yu2025dapo,
  title={Dapo: An open-source llm reinforcement learning system at scale},
  author={Yu, Qiying and Zhang, Zheng and Zhu, Ruofei and Yuan, Yufeng and Zuo, Xiaochen and Yue, Yu and Dai, Weinan and Fan, Tiantian and Liu, Gaohong and Liu, Lingjun and others},
  journal={arXiv preprint arXiv:2503.14476},
  year={2025}
}

@inproceedings{ahmadian2024back-leave-one-out,
  title={Back to Basics: Revisiting REINFORCE-Style Optimization for Learning from Human Feedback in LLMs},
  author={Ahmadian, Arash and Cremer, Chris and Gall{\'e}, Matthias and Fadaee, Marzieh and Kreutzer, Julia and Pietquin, Olivier and {\"U}st{\"u}n, Ahmet and Hooker, Sara},
  booktitle={Proceedings of the 62nd Annual Meeting of the Association for Computational Linguistics (Volume 1: Long Papers)},
  pages={12248--12267},
  year={2024}
}

@article{gehring2024rlef,
  title={Rlef: Grounding code llms in execution feedback with reinforcement learning},
  author={Gehring, Jonas and Zheng, Kunhao and Copet, Jade and Mella, Vegard and Carbonneaux, Quentin and Cohen, Taco and Synnaeve, Gabriel},
  journal={arXiv preprint arXiv:2410.02089},
  year={2024}
}

@inproceedings{
    jain2025multiturnthrougOne,
    title={Multi-Turn Code Generation Through Single-Step Rewards},
    author={Arnav Kumar Jain and Gonzalo Gonzalez-Pumariega and Wayne Chen and Alexander M Rush and Wenting Zhao and Sanjiban Choudhury},
    booktitle={Forty-second International Conference on Machine Learning},
    year={2025},
    url={https://openreview.net/forum?id=aJeLhLcsh0}
  }

@article{chen2025R1-Code-Interpreter,
  title={R1-Code-Interpreter: Training LLMs to Reason with Code via Supervised and Reinforcement Learning},
  author={Chen, Yongchao and Liu, Yueying and Zhou, Junwei and Hao, Yilun and Wang, Jingquan and Zhang, Yang and Fan, Chuchu},
  journal={arXiv preprint arXiv:2505.21668},
  year={2025}
}

@article{lambert2024tulu-RLVR,
  title={Tulu 3: Pushing frontiers in open language model post-training},
  author={Lambert, Nathan and Morrison, Jacob and Pyatkin, Valentina and Huang, Shengyi and Ivison, Hamish and Brahman, Faeze and Miranda, Lester James V and Liu, Alisa and Dziri, Nouha and Lyu, Shane and others},
  journal={arXiv preprint arXiv:2411.15124},
  year={2024}
}

@inproceedings{zeng-etal-2025-acecoder,
    title = "{ACECODER}: Acing Coder {RL} via Automated Test-Case Synthesis",
    author = "Zeng, Huaye  and
      Jiang, Dongfu  and
      Wang, Haozhe  and
      Nie, Ping  and
      Chen, Xiaotong  and
      Chen, Wenhu",
    editor = "Che, Wanxiang  and
      Nabende, Joyce  and
      Shutova, Ekaterina  and
      Pilehvar, Mohammad Taher",
    booktitle = "Proceedings of the 63rd Annual Meeting of the Association for Computational Linguistics (Volume 1: Long Papers)",
    month = jul,
    year = "2025",
    address = "Vienna, Austria",
    publisher = "Association for Computational Linguistics",
    url = "https://aclanthology.org/2025.acl-long.587/",
    doi = "10.18653/v1/2025.acl-long.587",
    pages = "12023--12040"
}

@article{dai2024processreward,
  title={Process supervision-guided policy optimization for code generation},
  author={Dai, Ning and Wu, Zheng and Zheng, Renjie and Wei, Ziyun and Shi, Wenlei and Jin, Xing and Liu, Guanlin and Dun, Chen and Huang, Liang and Yan, Lin},
  journal={arXiv preprint arXiv:2410.17621},
  year={2024}
}

@inproceedings{ye2025process,
  title={Process-supervised reinforcement learning for code generation},
  author={Ye, Yufan and Zhang, Ting and Jiang, Wenbin and Huang, Hua},
  booktitle={Proceedings of the 2025 Conference on Empirical Methods in Natural Language Processing},
  pages={14224--14237},
  year={2025}
}

@inproceedings{liu2025understandingr1-zero-likeTraining,
  title={Understanding r1-zero-like training: A critical perspective},
  author={Liu, Zichen and Chen, Changyu and Li, Wenjun and Qi, Penghui and Pang, Tianyu and Du, Chao and Lee, Wee Sun and Lin, Min},
  booktitle={Conference on Language Modeling (COLM)},
  year={2025}
}

@article{chen2021humaneval,
  title={Evaluating large language models trained on code},
  author={Chen, Mark and Tworek, Jerry and Jun, Heewoo and Yuan, Qiming and Pinto, Henrique Ponde De Oliveira and Kaplan, Jared and Edwards, Harri and Burda, Yuri and Joseph, Nicholas and Brockman, Greg and others},
  journal={arXiv preprint arXiv:2107.03374},
  year={2021}
}

@inproceedings{sheng2025hybridflow-verl,
  title={Hybridflow: A flexible and efficient rlhf framework},
  author={Sheng, Guangming and Zhang, Chi and Ye, Zilingfeng and Wu, Xibin and Zhang, Wang and Zhang, Ru and Peng, Yanghua and Lin, Haibin and Wu, Chuan},
  booktitle={Proceedings of the Twentieth European Conference on Computer Systems},
  pages={1279--1297},
  year={2025}
}

@article{liu2023humanevalplus,
  title={Is your code generated by chatgpt really correct? rigorous evaluation of large language models for code generation},
  author={Liu, Jiawei and Xia, Chunqiu Steven and Wang, Yuyao and Zhang, Lingming},
  journal={Advances in Neural Information Processing Systems},
  volume={36},
  pages={21558--21572},
  year={2023}
}

@article{zhuo2024bigcodebench,
  title={Bigcodebench: Benchmarking code generation with diverse function calls and complex instructions},
  author={Zhuo, Terry Yue and Vu, Minh Chien and Chim, Jenny and Hu, Han and Yu, Wenhao and Widyasari, Ratnadira and Yusuf, Imam Nur Bani and Zhan, Haolan and He, Junda and Paul, Indraneil and others},
  journal={arXiv preprint arXiv:2406.15877},
  year={2024}
}

@article{jain2024livecodebench,
  title={Livecodebench: Holistic and contamination free evaluation of large language models for code},
  author={Jain, Naman and Han, King and Gu, Alex and Li, Wen-Ding and Yan, Fanjia and Zhang, Tianjun and Wang, Sida and Solar-Lezama, Armando and Sen, Koushik and Stoica, Ion},
  journal={arXiv preprint arXiv:2403.07974},
  year={2024}
}

@article{quan2025codeelo,
  title={CodeElo: Benchmarking Competition-level Code Generation of LLMs with Human-comparable Elo Ratings},
  author={Quan, Shanghaoran and Yang, Jiaxi and Yu, Bowen and Zheng, Bo and Liu, Dayiheng and Yang, An and Ren, Xuancheng and Gao, Bofei and Miao, Yibo and Feng, Yunlong and others},
  journal={arXiv preprint arXiv:2501.01257},
  year={2025}
}

@article{li2023taco,
  title={TACO: Topics in Algorithmic COde generation dataset},
  author={Rongao Li and Jie Fu and Bo-Wen Zhang and Tao Huang and Zhihong Sun and Chen Lyu and Guang Liu and Zhi Jin and Ge Li},
  journal={arXiv preprint arXiv:2312.14852},
  year={2023}
}

@article{zhang2025reward-hacking,
  title={Chasing the Tail: Effective Rubric-based Reward Modeling for Large Language Model Post-Training},
  author={Zhang, Junkai and Wang, Zihao and Gui, Lin and Sathyendra, Swarnashree Mysore and Jeong, Jaehwan and Veitch, Victor and Wang, Wei and He, Yunzhong and Liu, Bing and Jin, Lifeng},
  journal={arXiv preprint arXiv:2509.21500},
  year={2025}
}

@article{schulman2017ppo,
  title={Proximal policy optimization algorithms},
  author={Schulman, John and Wolski, Filip and Dhariwal, Prafulla and Radford, Alec and Klimov, Oleg},
  journal={arXiv preprint arXiv:1707.06347},
  year={2017}
}

@article{kool2019buy-leave-one-out,
  title={Buy 4 REINFORCE Samples, Get a Baseline for Free!},
  author={Kool, Wouter and van Hoof, Herke and Welling, Max},
  booktitle={ICLR 2019 Deep Reinforcement Learning meets Structured Prediction Workshop},
  year={2019}
}

@inproceedings{feng2025groupingroup,
 author = {Feng, Lang and Xue, Zhenghai and Liu, Tingcong and An, Bo},
 booktitle = {Advances in Neural Information Processing Systems},
 editor = {D. Belgrave and C. Zhang and H. Lin and R. Pascanu and P. Koniusz and M. Ghassemi and N. Chen},
 pages = {46375--46408},
 publisher = {Curran Associates, Inc.},
 title = {Group-in-Group Policy Optimization for LLM Agent Training},
 url = {https://proceedings.neurips.cc/paper_files/paper/2025/file/420c9f777c0b4f78d515e53cf74d58b2-Paper-Conference.pdf},
 volume = {38},
 year = {2025}
}

@article{shojaee2023execution-ppocoder,
  title={Execution-based code generation using deep reinforcement learning},
  author={Shojaee, Parshin and Jain, Aneesh and Tipirneni, Sindhu and Reddy, Chandan K},
  journal={arXiv preprint arXiv:2301.13816},
  year={2023}
}

@article{roziere2023code-single-turn-code,
  title={Code llama: Open foundation models for code},
  author={Roziere, Baptiste and Gehring, Jonas and Gloeckle, Fabian and Sootla, Sten and Gat, Itai and Tan, Xiaoqing Ellen and Adi, Yossi and Liu, Jingyu and Sauvestre, Romain and Remez, Tal and others},
  journal={arXiv preprint arXiv:2308.12950},
  year={2023}
}

@article{wang2021codet5-single-turn,
  title={Codet5: Identifier-aware unified pre-trained encoder-decoder models for code understanding and generation},
  author={Wang, Yue and Wang, Weishi and Joty, Shafiq and Hoi, Steven CH},
  journal={arXiv preprint arXiv:2109.00859},
  year={2021}
}

@article{dong2025survey-multi-turn-code,
  title={A survey on code generation with llm-based agents},
  author={Dong, Yihong and Jiang, Xue and Qian, Jiaru and Wang, Tian and Zhang, Kechi and Jin, Zhi and Li, Ge},
  journal={arXiv preprint arXiv:2508.00083},
  year={2025}
}

@inproceedings{zheng2024opencodeinterpreter-multi-turn-code,
  title={OpenCodeInterpreter: Integrating Code Generation with Execution and Refinement},
  author={Zheng, Tianyu and Zhang, Ge and Shen, Tianhao and Liu, Xueling and Lin, Bill Yuchen and Fu, Jie and Chen, Wenhu and Yue, Xiang},
  booktitle={Findings of the Association for Computational Linguistics ACL 2024},
  pages={12834--12859},
  year={2024}
}

@article{williams1992Reinforce-algo,
  title={Simple statistical gradient-following algorithms for connectionist reinforcement learning},
  author={Williams, Ronald J},
  journal={Machine learning},
  volume={8},
  number={3},
  pages={229--256},
  year={1992},
  publisher={Springer}
}

@article{sun2025large-rewarddesign,
  title={A large language model-driven reward design framework via dynamic feedback for reinforcement learning},
  author={Sun, Shengjie and Liu, Runze and Lyu, Jiafei and Yang, Jing-Wen and Zhang, Liangpeng and Li, Xiu},
  journal={Knowledge-Based Systems},
  volume={326},
  pages={114065},
  year={2025},
  publisher={Elsevier}
}

@article{xu2025policy-rewarddesign,
  title={The Policy Cliff: A Theoretical Analysis of Reward-Policy Maps in Large Language Models},
  author={Xu, Xingcheng},
  journal={arXiv preprint arXiv:2507.20150},
  year={2025}
}

@inproceedings{jimenez2024swe,
  title={Swe-bench: Can language models resolve real-world github issues?},
  author={Jimenez, Carlos E and Yang, John and Wettig, Alexander and Yao, Shunyu and Pei, Kexin and Press, Ofir and Narasimhan, Karthik},
  booktitle={International Conference on Learning Representations},
  volume={2024},
  pages={54107--54157},
  year={2024}
}

@article{yang2024swe,
  title={Swe-agent: Agent-computer interfaces enable automated software engineering},
  author={Yang, John and Jimenez, Carlos E and Wettig, Alexander and Lieret, Kilian and Yao, Shunyu and Narasimhan, Karthik and Press, Ofir},
  journal={Advances in Neural Information Processing Systems},
  volume={37},
  pages={50528--50652},
  year={2024}
}

@inproceedings{li2025codeprm,
  title={Codeprm: Execution feedback-enhanced process reward model for code generation},
  author={Li, Qingyao and Dai, Xinyi and Li, Xiangyang and Zhang, Weinan and Wang, Yasheng and Tang, Ruiming and Yu, Yong},
  booktitle={Findings of the Association for Computational Linguistics: ACL 2025},
  pages={8169--8182},
  year={2025}
}

@misc{penedo2025codeforces,
      title={CodeForces CoTs}, 
      author={Guilherme Penedo and Anton Lozhkov and Hynek Kydlíček and Loubna Ben Allal and Edward Beeching and Agustín Piqueres Lajarín and Quentin Gallouédec and Nathan Habib and Lewis Tunstall and Leandro von Werra},
      year={2025},
      publisher = {Hugging Face},
      journal = {Hugging Face repository},
      howpublished = {\url{https://huggingface.co/datasets/open-r1/codeforces-cots}}
}

@misc{xia2025leetcodedatasettemporaldatasetrobust,
      title={LeetCodeDataset: A Temporal Dataset for Robust Evaluation and Efficient Training of Code LLMs}, 
      author={Yunhui Xia and Wei Shen and Yan Wang and Jason Klein Liu and Huifeng Sun and Siyue Wu and Jian Hu and Xiaolong Xu},
      year={2025},
      eprint={2504.14655},
      archivePrefix={arXiv},
      primaryClass={cs.LG},
      url={https://arxiv.org/abs/2504.14655}, 
}

@article{li2022competition-alphacode,
  title={Competition-level code generation with alphacode},
  author={Li, Yujia and Choi, David and Chung, Junyoung and Kushman, Nate and Schrittwieser, Julian and Leblond, R{\'e}mi and Eccles, Tom and Keeling, James and Gimeno, Felix and Dal Lago, Agustin and others},
  journal={Science},
  volume={378},
  number={6624},
  pages={1092--1097},
  year={2022},
  publisher={American Association for the Advancement of Science}
}

@article{wang2024enhancing-rlwithcode,
  title={Enhancing code llms with reinforcement learning in code generation: A survey},
  author={Wang, Junqiao and Zhang, Zeng and He, Yangfan and Zhang, Zihao and Song, Xinyuan and Song, Yuyang and Shi, Tianyu and Li, Yuchen and Xu, Hengyuan and Wu, Kunyu and others},
  journal={arXiv preprint arXiv:2412.20367},
  year={2024}
}

@misc{zhang2025surveyreinforcementlearninglarge-rlwithcode,
      title={A Survey of Reinforcement Learning for Large Reasoning Models}, 
      author={Kaiyan Zhang and Yuxin Zuo and Bingxiang He and Youbang Sun and Runze Liu and Che Jiang and Yuchen Fan and Kai Tian and Guoli Jia and Pengfei Li and Yu Fu and Xingtai Lv and Yuchen Zhang and Sihang Zeng and Shang Qu and Haozhan Li and Shijie Wang and Yuru Wang and Xinwei Long and Fangfu Liu and Xiang Xu and Jiaze Ma and Xuekai Zhu and Ermo Hua and Yihao Liu and Zonglin Li and Huayu Chen and Xiaoye Qu and Yafu Li and Weize Chen and Zhenzhao Yuan and Junqi Gao and Dong Li and Zhiyuan Ma and Ganqu Cui and Zhiyuan Liu and Biqing Qi and Ning Ding and Bowen Zhou},
      year={2025},
      eprint={2509.08827},
      archivePrefix={arXiv},
      primaryClass={cs.CL}, 
}

@article{liu2023rltf,
  title={Rltf: Reinforcement learning from unit test feedback},
  author={Liu, Jiate and Zhu, Yiqin and Xiao, Kaiwen and Fu, Qiang and Han, Xiao and Yang, Wei and Ye, Deheng},
  journal={arXiv preprint arXiv:2307.04349},
  year={2023}
}

\appendix

\section{Dataset-Level Test-Suite Statistics}\label{code_dataset_statistics}

Table~\ref{tab:code_dataset_statistics} reports the number of test cases per
problem across representative code-generation benchmarks. These datasets
consistently contain multiple test cases per problem, supporting our setting
where partial success can be observed directly from test-case-level execution
feedback.
\begin{table}[h]
\centering
\footnotesize
\setlength{\tabcolsep}{1.5pt}
\begin{tabular}{lccc}
\toprule
Dataset & Total Examples & Mean Tests & Median Tests \\
\midrule
CodeContests    & 13,328 & 95.91  & 101 \\
LeetCodeDataset & 2,641  & 100.10 & 99 \\
codeforces-cots   & 9,556  & 51.07  & 43 \\
TACO                 & 25,443 & 53.36  & 10 \\
PrimeIntellect & 16,252 & 89.67 & 101\\
\bottomrule
\end{tabular}
\caption{Dataset-level test-suite statistics for five representative code-generation datasets: deepmind/code\_contests \cite{li2022competition-alphacode}, newfacade/LeetCodeDataset \cite{xia2025leetcodedatasettemporaldatasetrobust}, open-r1/codeforces-cots \cite{penedo2025codeforces}, BAAI/TACO \cite{li2023taco} and the PrimeIntellect subset from agentica-org/DeepCoder-Preview-Dataset \cite{luo2025deepcoder}.}\label{tab:code_dataset_statistics}
\end{table}

\begin{figure}[th]
  \centering

  \begin{subfigure}{\columnwidth}
    \centering
    \includegraphics[width=\columnwidth]{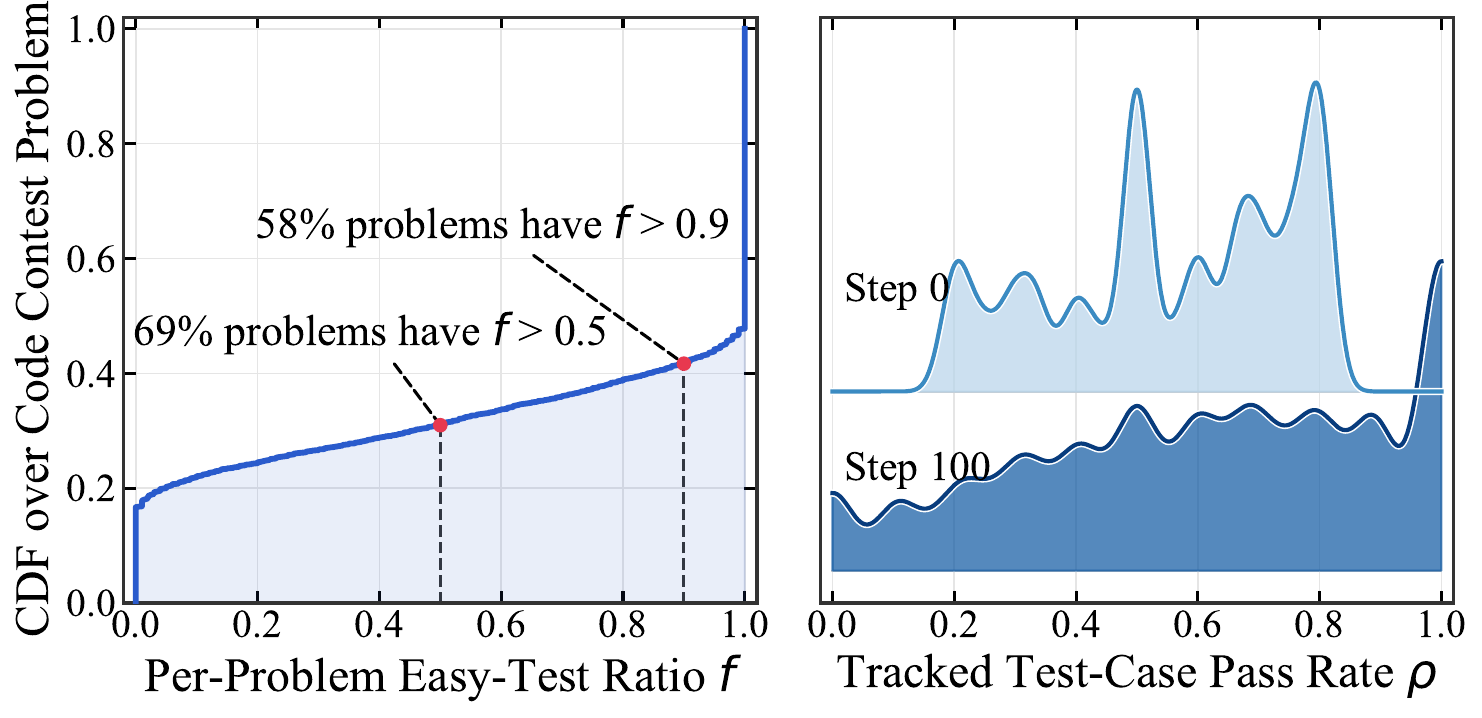}
    \caption{\textsc{CodeContests}.}
    \label{fig:app_codecontests_skew}
  \end{subfigure}

  \vspace{0.6em}

  \begin{subfigure}{\columnwidth}
    \centering
    \includegraphics[width=\columnwidth]{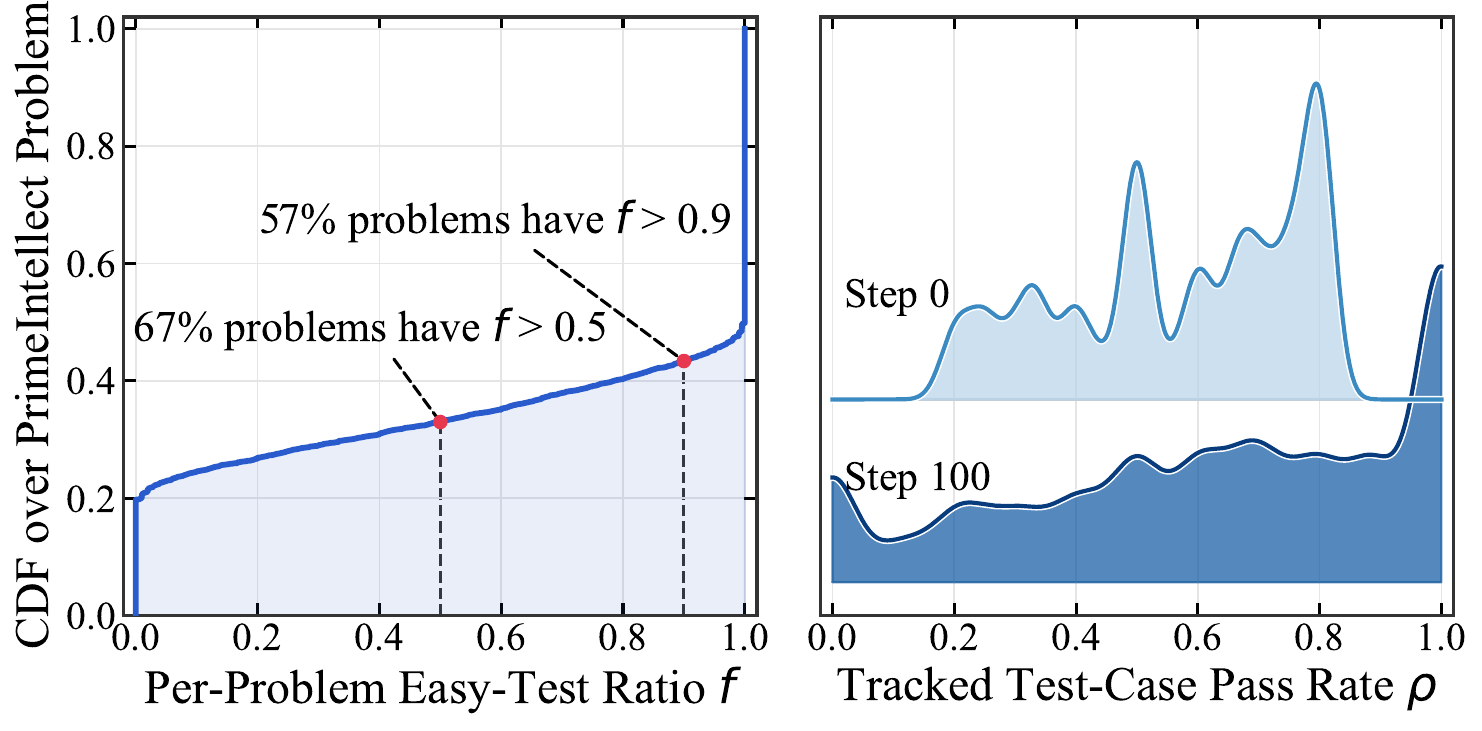}
    \caption{\textsc{PrimeIntellect}.}
    \label{fig:app_primeintellect_skew}
  \end{subfigure}

  \caption{
  Additional dataset-level evidence of test-case difficulty skew. 
  For each dataset, the left panel plots the CDF of the per-problem easy-test ratio ($f$), illustrating the abundance of initially easy test cases. 
  The right panel tracks the pass-rate distribution shifts of initially intermediate test cases ($\rho\in[0.2,0.8]$) across training steps.
  }
  \label{fig:app_additional_skew}
\end{figure}

\section{Additional Evidence of Test-Case Difficulty Skew}
\label{app:additional_skew}
To show that test-case difficulty skew is a general phenomenon in code generation rather than an artifact of a particular benchmark, we extend the same analysis to two additional competitive code-generation datasets: \textsc{CodeContests} (deepmind/code\_contests~\cite{li2022competition-alphacode}) and \textsc{PrimeIntellect} (for the latter, we randomly sample 10,000 examples from the PrimeIntellect subset in agentica-org/DeepCoder-Preview-Dataset~\cite{luo2025deepcoder} with seed 0). Following the analysis protocol of Fig.~\ref{fig:method1}, we define the per-problem easy-test ratio $f$ as the fraction of test cases in problem $x$ whose empirical pass rate satisfies $\rho>0.8$. We also track the pass-rate distribution of test cases that are initially in the intermediate regime, i.e., $\rho\in[0.2,0.8]$ at Step 0.

Fig.~\ref{fig:app_additional_skew} shows that the same pattern persists across datasets. Specifically, a large fraction of problems contain many high-pass-rate test cases before training: on \textsc{CodeContests}, 69\% of problems have $f>0.5$ and 58\% have $f>0.9$; similarly, on \textsc{PrimeIntellect}, 67\% of problems have $f>0.5$ and 57\% have $f>0.9$. Furthermore, initially intermediate test cases tend to migrate toward higher pass-rate regions as training proceeds. These consistency results provide additional evidence that the partial-success signal in code generation is structurally skewed toward already-easy test cases, rather than being uniformly distributed over tests of different learning value.

\section{Case Study: Partial-Success Biases}
\label{app:case_study}

This section presents two rollout-level case studies comparing
test-suite-level outcome rewards, difficulty-weighted partial-success
rewards (Diff), and VeRPO.
Following Section~\ref{VeRPO}, Diff uses the
uncalibrated difficulty-aware weight
\(w_j=\exp(-\alpha\rho_j)\), while VeRPO uses the density-calibrated
weight
\(w'_j=\exp(-\alpha\rho_j)/(\hat N(\rho_j)+\delta)\).
We set \(\alpha=2\) for both Diff and VeRPO, and use the adaptive KDE
bandwidth
\(\sigma=\operatorname{std}(\{\rho_j\}_{j=1}^{|\mathcal{U}_x|})/2\).
The outcome-driven reward is 1 if all tests pass and 0 otherwise. Candidate A and
Candidate B are two representative members of the same group.

\subsection{Correcting Easy-Test Cardinality Bias}
\label{app:case_easy_cardinality}

Consider a rollout group where the test suite contains many already easy
tests and only a few frontier tests. Candidate A passes many easy tests
but none of the frontier tests, whereas Candidate B passes fewer easy
tests but solves several frontier tests. Since both candidates still fail
the complete test suite, outcome-driven rewards cannot distinguish them.
Table~\ref{tab:case_easy_cardinality} shows a concrete toy instance with
100 easy tests with empirical pass rate \(\rho=0.95\) and 10 frontier
tests with \(\rho=0.20\).  Candidate A passes 90 easy tests and 0
frontier tests, while Candidate B passes 50 easy tests and 5 frontier
tests.

\begin{center}
\centering
\footnotesize
\setlength{\tabcolsep}{1.6pt}
\begin{tabular}{lcccccc}
\toprule
Set & \# & \(\rho\) & A pass & B pass & \(w_j\) (Diff) & \(w'_j\) (VeRPO) \\
\midrule
Easy & 100 & 0.95 & 90 & 50 & 0.150 & 0.0015 \\
Frontier & 10 & 0.20 & 0 & 5 & 0.670 & 0.067 \\
\bottomrule
\end{tabular}

\vspace{0.25em}

\setlength{\tabcolsep}{2.6pt}
\begin{tabular}{lcc}
\toprule
Reward & \(R(A)\) & \(R(B)\) \\
\midrule
Outcome-driven & 0.00 & 0.00 \\
Diff & 13.50 & 10.85 \\
VeRPO & 0.14 & 0.41 \\
\bottomrule
\end{tabular}
{\captionsetup{type=table,hypcap=false}
\caption{Case 1 setup and resulting rewards. Weights are test-level and
shared across candidates.}
\label{tab:case_easy_cardinality}
}
\end{center}

This example illustrates the cardinality bias analyzed in
Section~\ref{Partial Success and Cardinality Bias}. Although Diff assigns
smaller per-test weights to easy tests, their large quantity still makes
Candidate A receive a higher reward than Candidate B. VeRPO instead
normalizes test contributions by the local empirical density
\(\hat N(\rho_j)\), reducing the aggregate dominance of easy tests and
correctly assigning a higher reward to the candidate that makes frontier
progress.
\subsection{Amplifying Sparse Frontier Progress}
\label{app:case_frontier_progress}

The second case studies a subtler failure mode. Candidate A and
Candidate B pass the same large set of easy tests, but Candidate B
additionally passes a frontier test. Both candidates still fail the
complete suite, so outcome-driven rewards again provide no distinction.
Although Diff increases Candidate B's raw reward, the group-mean baseline
can still be dominated by the easy-test mass, yielding a weak or even
negative relative advantage.

Table~\ref{tab:case_frontier_progress} shows an illustrative group with
100 easy tests of empirical pass rate \(\rho=0.95\) and one frontier test
of empirical pass rate \(\rho=0.20\). Candidate A and Candidate B both
pass 90 easy tests, while Candidate B additionally passes the frontier
test.

\begin{center}
\centering
\footnotesize
\setlength{\tabcolsep}{1.6pt}
\begin{tabular}{lcccccc}
\toprule
Set & \# & \(\rho\) & A pass & B pass & \(w_j\) (Diff) & \(w'_j\) (VeRPO) \\
\midrule
Easy & 100 & 0.95 & 90 & 90 & 0.150 & 0.0015 \\
Frontier & 1 & 0.20 & 0 & 1 & 0.670 & 0.670 \\
\bottomrule
\end{tabular}

\vspace{0.25em}

\setlength{\tabcolsep}{1.6pt}
\begin{tabular}{lccccc}
\toprule
Reward & \(R(A)\) & \(R(B)\) & \(\bar{R}\) & \(\mathrm{Adv}(A)\) & \(\mathrm{Adv}(B)\) \\
\midrule
Outcome-driven & 0.00 & 0.00 & 0.00 & 0.00 & 0.00 \\
Diff & 13.50 & 14.17 & 14.38 & -0.88 & -0.21 \\
VeRPO & 0.14 & 0.81 & 0.28 & -0.14 & 0.53 \\
\bottomrule
\end{tabular}
{\captionsetup{type=table,hypcap=false}
\caption{Case 2 setup and resulting advantages. The group mean $\bar R$ is computed over the full group. Weights are test-level and shared across candidates.}
\label{tab:case_frontier_progress}
}
\end{center}

This case shows that correct reward ranking alone is insufficient for
group-based RL; what matters is the relative advantage after subtracting
the group-mean baseline. Under Diff, the easy-test mass inflates
\(\bar r^{\mathrm{turn}}\). Candidate B thus still receives a negative advantage despite passing the frontier test. VeRPO reduces the
common easy-test background through density calibration, turning the same
frontier progress into a positive and substantially stronger advantage.

\section{Detailed Proofs}\label{sec:appendix}

\subsection{Derivation of the Integral Form in Eq.~\eqref{eq:density_form}}\label{Derivation_density_form}

In this section, we provide the rigorous derivation to transform the discrete summation $\bar{r}^{\,\text{turn}} = \sum_{j=1}^{|\mathcal{U}_x|} w_j \rho_j$ into the integral form. 

Since multiple test cases may share the same empirical pass rate $\rho$ but possess different weights $w_j$, we first group the summation by the unique values of $\rho$. Let $\mathcal{V}_\rho \subset [0, 1]$ be the set of all unique empirical pass rates in $\mathcal{U}_x$. For a given pass rate $\rho \in \mathcal{V}_\rho$, we define its corresponding index subset as:
\begin{equation}
    \mathcal{I}(\rho) = \{ j \in \{1, \dots, |\mathcal{U}_x|\} \mid \rho_j = \rho \}.
\end{equation}
Let $N_\rho = |\mathcal{I}(\rho)|$ be the number of test cases with pass rate $\rho$. We define the conditional average weight for the subset $\mathcal{I}(\rho)$ as:
\begin{equation}
    \bar{w}(\rho) = \frac{1}{N_\rho} \sum_{j \in \mathcal{I}(\rho)} w_j.
\end{equation}
Thus, the sum of weights for a specific pass rate $\rho$ is equivalent to $N_\rho \bar{w}(\rho)$. We can rewrite Eq.~\eqref{eq:baseline_2} by summing over the unique values in $\mathcal{V}_\rho$:
\begin{equation}
    \sum_{j=1}^{|\mathcal{U}_x|} w_j \rho_j = \sum_{\rho \in \mathcal{V}_\rho} \left( \sum_{j \in \mathcal{I}(\rho)} w_j \right) \rho = \sum_{\rho \in \mathcal{V}_\rho} \bar{w}(\rho) N_\rho \rho.
    \label{eq:discrete_grouped}
\end{equation}

To generalize this into an integral over the continuous interval $[0, 1]$, we define the unnormalized empirical test density using the Dirac delta function $\delta(\cdot)$:
\begin{equation}
    N(\rho) = \sum_{j=1}^{|\mathcal{U}_x|} \delta(\rho - \rho_j).
\end{equation}
Notice that integrating $N(\rho)$ over an infinitesimally small interval containing $\rho^* \in \mathcal{V}_\rho$ yields exactly the count $N_{\rho^*}$. Using the sifting property of the Dirac delta function, we evaluate the integral:
\begin{equation}
    \int_0^1 \bar{w}(\rho)\,\rho\,N(\rho)\, d\rho = \int_0^1 \bar{w}(\rho)\,\rho \sum_{j=1}^{|\mathcal{U}_x|} \delta(\rho - \rho_j)\, d\rho.
\end{equation}
Interchanging the integral and the summation gives:
\begin{equation}
    \sum_{j=1}^{|\mathcal{U}_x|} \int_0^1 \bar{w}(\rho)\,\rho\, \delta(\rho - \rho_j)\, d\rho = \sum_{j=1}^{|\mathcal{U}_x|} \bar{w}(\rho_j) \rho_j.
\end{equation}
Grouping this final summation back into the unique subsets $\mathcal{V}_\rho$, we obtain:
\begin{equation}
\begin{aligned}
\sum_{\rho^* \in \mathcal{V}_\rho} \sum_{j \in \mathcal{I}(\rho^*)} \bar{w}(\rho^*) \rho^* 
&= \sum_{\rho^* \in \mathcal{V}_\rho} N_{\rho^*} \bar{w}(\rho^*) \rho^* \notag\\
&= 
\sum_{\rho^* \in \mathcal{V}_\rho} \sum_{j \in \mathcal{I}(\rho^*)} w_j \rho^* \notag\\ &= 
\sum_{j=1}^{|\mathcal{U}_x|} w_j \rho_j.
\end{aligned}
\end{equation}
This completes the derivation, proving that $\bar{r}^{\,\text{turn}} = \int_0^1 \bar{w}(\rho)\,\rho\,N(\rho)\, d\rho$ is mathematically equivalent to Eq.~\eqref{eq:baseline_2}.

\subsection{Formal Analysis of the Cardinality Effect}
\label{app:contribution_gap_bound}

Building upon the integral form derived in Eq.~\eqref{eq:density_form}, we partition the test cases by a mastery threshold \(\epsilon \in (0,1)\). Specifically, the discrete sets for the conquered tests \(\mathcal U_{\mathcal C} = \{u_j \in \mathcal U_x : \epsilon \le \rho_j \le 1\}\) and the frontier tests \(\mathcal U_{\mathcal F} = \{u_j \in \mathcal U_x : 0 \le \rho_j < \epsilon\}\) correspond to the continuous integration domains \([\epsilon, 1]\) and \([0, \epsilon)\) over the empirical density \(N(\rho)\), respectively.

Let the cardinalities of the two sets be defined via the density function as:
\[
N_{\mathcal C} = \int_\epsilon^1 N(\rho)\, d\rho, \quad
N_{\mathcal F} = \int_0^\epsilon N(\rho)\, d\rho.
\]
Accordingly, their respective contributions to the group-mean baseline can be expressed as:
\[
\begin{split}
R_{\mathcal C} = \int_\epsilon^1 \bar{w}(\rho)\,\rho\,N(\rho)\, d\rho, \\
R_{\mathcal F} = \int_0^\epsilon \bar{w}(\rho)\,\rho\,N(\rho)\, d\rho.
\end{split}
\]

\begin{theorem}[Contribution-Gap Bound over Test Subsets]
\label{tho:cardinality_bias_appendix}
Assume the test weights satisfy \(0 < W_{\min} \le w_j \le W_{\max} < \infty\) for all \(j\in \{1,\dots,|\mathcal U_x|\}\), which implies \(W_{\min} \le \bar{w}(\rho) \le W_{\max}\) for any \(\rho\) where \(N(\rho) > 0\). Let \(\Delta R = R_{\mathcal{C}} - R_{\mathcal{F}}\) denote the contribution gap between the conquered test set and the frontier test set. Then,
\begin{equation}
\epsilon\bigl(W_{\min}N_{\mathcal C}-W_{\max}N_{\mathcal F}\bigr)
\;\le\; \Delta R \;\le\;
W_{\max}N_{\mathcal C}.
\label{eq:region_gap_appendix}
\end{equation}
\end{theorem}

\begin{proof}
We first establish the lower bound for \(\Delta R\) by bounding \(R_{\mathcal C}\) and \(R_{\mathcal F}\) separately.
For \(R_{\mathcal C}\), the integration domain is \([\epsilon, 1]\). Since \(\rho \ge \epsilon\), \(\bar{w}(\rho) \ge W_{\min}\) over this domain, and \(N(\rho) \ge 0\), we have:
\begin{equation}
\begin{aligned}
R_{\mathcal C} 
&= \int_\epsilon^1 \bar{w}(\rho)\,\rho\,N(\rho)\, d\rho \\
&\ge \int_\epsilon^1 W_{\min}\,\epsilon\,N(\rho)\, d\rho \\
&= \epsilon W_{\min} \int_\epsilon^1 N(\rho)\, d\rho = \epsilon W_{\min} N_{\mathcal C}.
\end{aligned}
\label{eq:RC_lower_integral}
\end{equation}

For \(R_{\mathcal F}\), the integration domain is \([0, \epsilon)\). Here, \(\rho < \epsilon\) and \(\bar{w}(\rho) \le W_{\max}\). Thus:
\begin{equation}
\begin{aligned}
R_{\mathcal F} 
&= \int_0^\epsilon \bar{w}(\rho)\,\rho\,N(\rho)\, d\rho \\
&\le \int_0^\epsilon W_{\max}\,\epsilon\,N(\rho)\, d\rho \\
&= \epsilon W_{\max} \int_0^\epsilon N(\rho)\, d\rho = \epsilon W_{\max} N_{\mathcal F}.
\end{aligned}
\label{eq:RF_upper_integral}
\end{equation}
Subtracting Eq.~\eqref{eq:RF_upper_integral} from Eq.~\eqref{eq:RC_lower_integral} yields the lower bound:
\[
R_{\mathcal C} - R_{\mathcal F} \ge \epsilon\bigl(W_{\min}N_{\mathcal C} - W_{\max}N_{\mathcal F}\bigr).
\]

Next, we establish the upper bound. For \(R_{\mathcal C}\), since \(\rho \le 1\) on \([\epsilon, 1]\) and \(\bar{w}(\rho) \le W_{\max}\), we have:
\begin{equation}
\begin{aligned}
R_{\mathcal C} 
&= \int_\epsilon^1 \bar{w}(\rho)\,\rho\,N(\rho)\, d\rho \\
&\le \int_\epsilon^1 W_{\max} \cdot 1 \cdot N(\rho)\, d\rho \\
&= W_{\max} \int_\epsilon^1 N(\rho)\, d\rho = W_{\max} N_{\mathcal C}.
\end{aligned}
\label{eq:RC_upper_integral}
\end{equation}
Furthermore, since \(\bar{w}(\rho) \ge W_{\min} > 0\), \(\rho \ge 0\), and \(N(\rho) \ge 0\) over the domain \([0, \epsilon)\), the integral for the frontier set is non-negative:
\begin{equation}
R_{\mathcal F} = \int_0^\epsilon \bar{w}(\rho)\,\rho\,N(\rho)\, d\rho \ge 0.
\end{equation}
Therefore, we conclude:
\[
R_{\mathcal C} - R_{\mathcal F} \le W_{\max}N_{\mathcal C} - 0 = W_{\max}N_{\mathcal C}.
\]
This proves the upper bound of Eq.~\eqref{eq:region_gap_appendix} and completes the proof.
\end{proof}

Theorem~\ref{tho:cardinality_bias_appendix} shows that the contribution gap is explicitly controlled by the cardinalities $N_{\mathcal C}$ and $N_{\mathcal F}$. 
In the skewed regime observed in practice, where conquered tests are far more numerous than frontier tests, i.e., $N_{\mathcal C}\gg N_{\mathcal F}$, the contribution gap is governed primarily by the size of the conquered partition. 
This formalizes the cardinality effect: even when per-test weights are introduced, the optimization baseline can be dominated by the sheer number of already-conquered tests unless this density effect is explicitly calibrated.

\subsection{Full Optimization Objective of VeRPO}
\label{app:verpo_objective}

Given the unified turn-level advantage $A(\tau_i,t)$ defined in Section~\ref{method}, VeRPO optimizes the policy with a clipped group-based objective. 
\begin{equation}
\begin{split}
  \mathcal{J}_{\mathrm{VeRPO}}(\theta)
  = &\ \mathbb{E}_{\substack{x \sim \mathcal{X} \\
  \{\tau_i\}_{i=1}^N \sim \pi_{\theta_{\mathrm{old}}}}}
  \Biggl[
  \frac{1}{\sum_{i=1}^N |\tau_i|}
  \sum_{i=1}^N \sum_{t=1}^{|\tau_i|} \\
  &\hspace{-1.3em}\min \Biggl(
  \psi_\theta(y_t^{(i)}) A(\tau_i,t),
  \operatorname{clip}\Bigl(
  \psi_\theta(y_t^{(i)}),\\ &
  1-\epsilon_{\mathrm{low}},
  1+\epsilon_{\mathrm{high}}
  \Bigr) A(\tau_i,t)
  \Biggr)
  \Biggr].
\end{split}
\label{eq:verpo_objective}
\end{equation}
where
\(
\psi_\theta(y_t^{(i)})
=
\frac{
\pi_\theta(y_t^{(i)} \mid s_t^{(i)})
}{
\pi_{\theta_{\mathrm{old}}}(y_t^{(i)} \mid s_t^{(i)})
}
\)
is the importance sampling ratio, and $\epsilon_{\mathrm{low}}$ and $\epsilon_{\mathrm{high}}$ are clipping thresholds.

\section{Training and Evaluation Details}
\label{sec:appendix_training_evaluation_details}
All experiments in this paper are implemented on top of the veRL framework \cite{sheng2025hybridflow-verl} and conducted on 8× NVIDIA H800 GPUs. For fair comparison, all methods use the same training and evaluation configuration unless otherwise specified. Source code will be made publicly available upon acceptance. 
\paragraph{Training Configuration.}  We utilize a rollout batch size of 32 code problems, with 10 responses sampled per problem. We set the maximum response length to 16384, with sampling parameters temperature = 1.0, top-p = 1.0 and top-k = -1.0. The policy learning rate is fixed at $1 \times 10^{-6}$. We set clipping parameters $\epsilon_\text{low}$ = 0.2 and $\epsilon_\text{high}$ = 0.28. All RL methods omit the KL divergence penalty to foster exploration. For VeRPO, the difficulty sensitivity coefficient $\alpha$ is set to 2.0, the decay factor $\gamma$ is set to 0.95, and the advantage balancing coefficient $\beta$ is set to 1.0 without additional tuning.
\paragraph{Evaluation Configuration.} Following \cite{yang2025qwen3}, we set sampling parameters to temperature 0.6, top-p 0.95, and top-k 20. The maximum response length is set to 16384. To rigorously evaluate functional correctness, we adopt the unbiased estimator for pass@$k$ proposed by \citealp{chen2021humaneval}, which calculates the expected probability that at least one of the top-$k$ generated code solutions passes the unit tests, defined as:$$\text{pass}@k := \mathbb{E}_{\text{problems}} \left[ 1 - \frac{\binom{n-c}{k}}{\binom{n}{k}} \right],$$ where $n$ denotes the number of samples generated per problem, and $c$ is the number of samples that pass all unit tests. In our experiments, we generate $n=8$ candidate solutions for each problem to robustly estimate the pass@1 metric.

\section{Training Analysis}
\subsection{Training Stability Analysis}\label{sec:appendix_training_stability_analysis}
\begin{figure}[h]
    \centering
    \includegraphics[width=\linewidth]{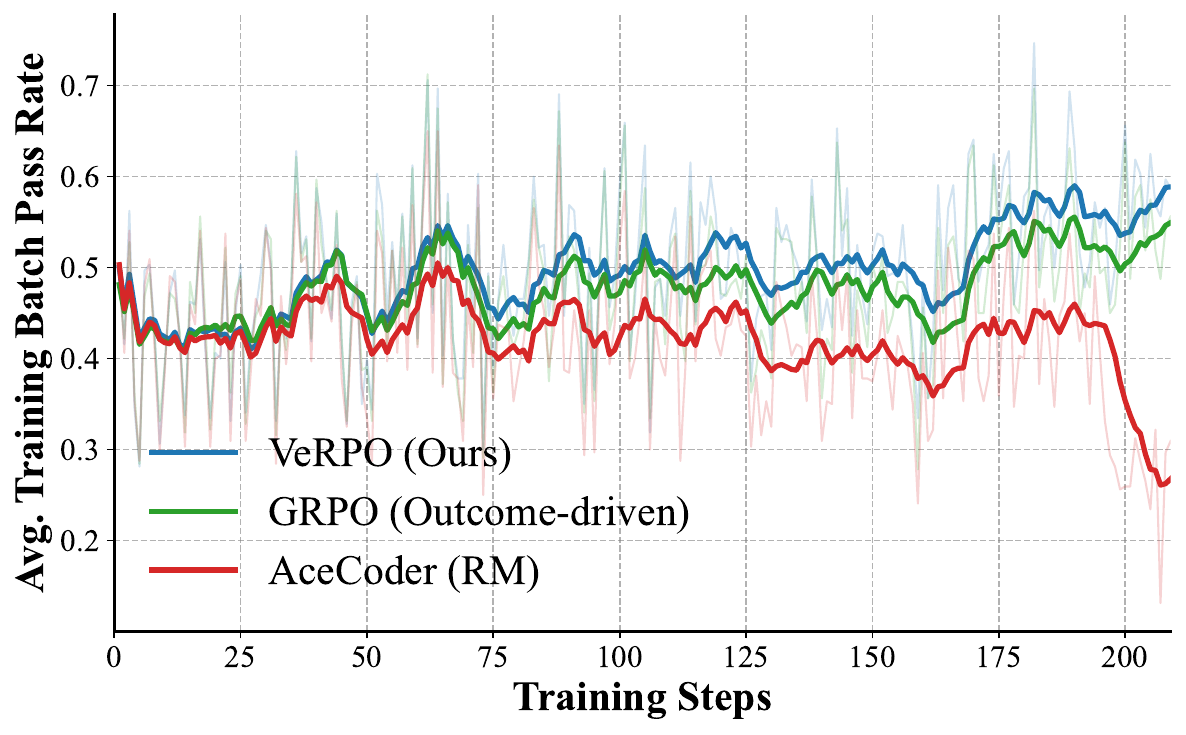}
    \caption{The average ground-truth pass rate computed on the online training rollout batches.
    }
    \label{fig:collapse_curve}
\end{figure}
In Section~\ref{Performance}, we observed that the RM-based baseline, AceCoder (MT), suffers from training collapse in multi-turn settings. Here, we provide a detailed visualization and analysis of this phenomenon.

The evolution of rollout performance is visualized in Figure~\ref{fig:collapse_curve}, which tracks the online average pass rate measured directly on the on-policy rollout batches.  While all methods show comparable performance in early stages, the RM-based AceCoder fails to exhibit a distinct upward trend and succumbs to catastrophic collapse after approximately 190 training steps. In contrast, VeRPO and the outcome-driven GRPO, which are both fully grounded in verifiable execution feedback, maintain robust stability and improvement. This phenomenon highlights a practical vulnerability of relying on fixed reward models for code generation optimization. Since it is impractical for learned RMs to exhaustively cover the infinite semantic space of complex programs, they inevitably yield misaligned signals for out-of-distribution responses, thereby destabilizing the optimization process.  Conversely, signals derived from verifiable execution feedback are inherently immune to such distribution shifts, further underscoring the necessity of VeRPO's design to generate dense learning signals fully grounded in execution feedback.

\subsection{Training Dynamics}
\begin{figure}[h]
    \centering
    \includegraphics[width=\linewidth]{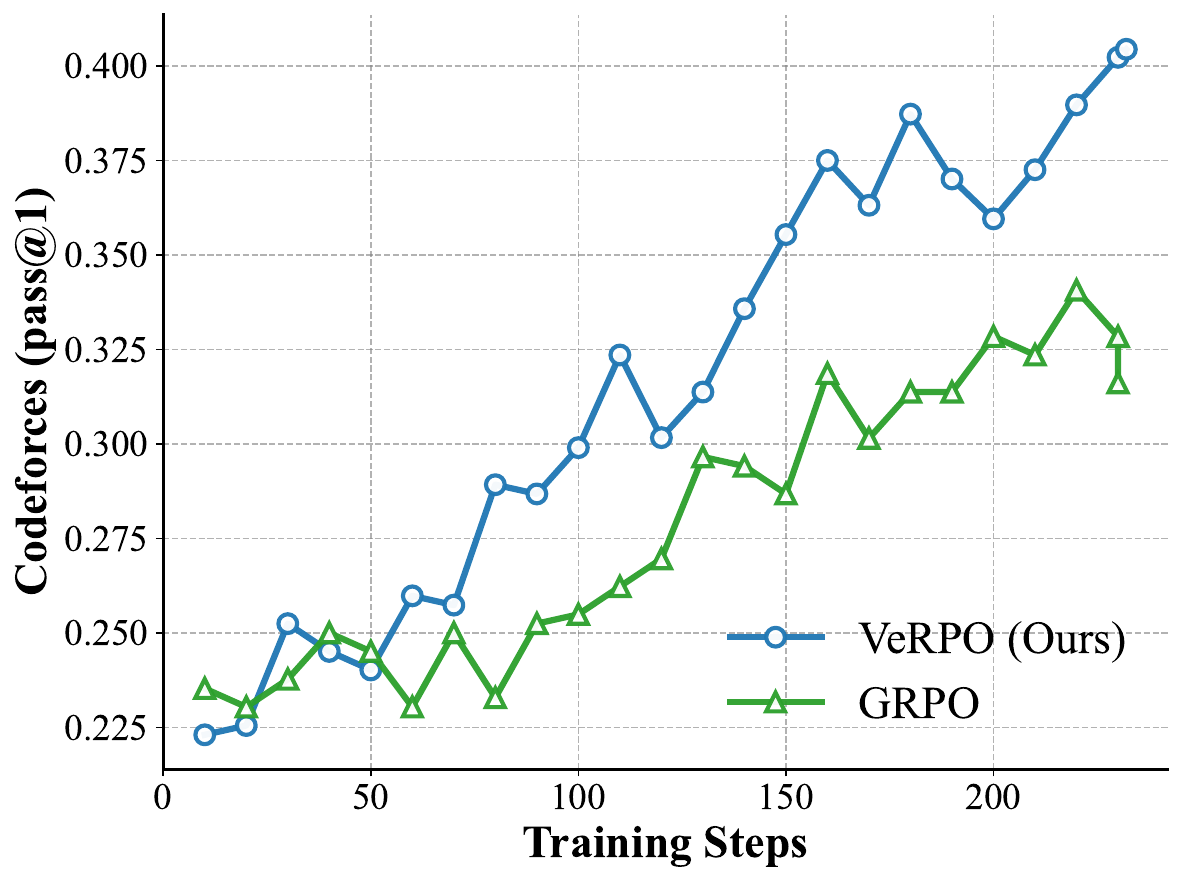}
    \caption{Training dynamics on Codeforces (pass@1).
    }
    \label{fig:training_dynamic}
\end{figure}
We analyze model evolution during training by tracking the performance on Codeforces from CodeElo~\cite{quan2025codeelo} using the unbiased pass@1 metric. 
As illustrated in  Fig.~\ref{fig:training_dynamic}, while both VeRPO (Ours) and GRPO exhibit comparable performance in the initial training phase (first 50 steps), VeRPO demonstrates significantly superior sample efficiency and convergence stability as training progresses. Starting from approximately Step 80, VeRPO establishes a distinct performance lead, maintaining a robust upward trend without the plateauing observed in the GRPO baseline.  Crucially, VeRPO achieves a peak pass@1 of $\sim$40.0\% at the end of training,  surpassing GRPO ($\sim$34.0\%) by a substantial margin. 
This divergence highlights that VeRPO facilitates more effective policy optimization, enabling the model to explore and stabilize on higher-quality solutions given the same computational budget.

\begin{table}[h]
\centering
\fontsize{8.5pt}{9.5pt}\selectfont
\setlength{\tabcolsep}{1.15pt}
\begin{tabular}{lccccc} 
\toprule
\multirow{2}{*}{Method} & HumanEval & BigCodeBench & LCB & Codeforces & \multirow{2}{*}{Avg} \\ 
\cmidrule(lr){2-2} \cmidrule(lr){3-3} \cmidrule(lr){4-4} \cmidrule(lr){5-5}
 & Plus & Full & V6 & CodeElo \\ 
\midrule
Qwen3-4B & 83.91 & 49.76 & 20.46 & 25.57 & 44.93 \\
\midrule
GRPO & 90.32 & 59.30 & 30.05 & 29.64 & 52.33\\
VeRPO & \textbf{92.84} & \textbf{61.76} & \textbf{36.34} & \textbf{32.50} & \textbf{55.86} \\ 
\bottomrule
\end{tabular}
\caption{Additional main results with Qwen3-4B. All values are pass@1 scores.}
\label{tab:qwen3_4b_main_results}
\end{table}

\begin{table}[ht]
\fontsize{10pt}{9.9pt}\selectfont
\renewcommand{\arraystretch}{1.3}
    \centering
    \begin{tabular}{cc}
        \toprule
        \textbf{Metric} & \textbf{Statistics} \\
        \midrule
        \# Total Examples & $7436$ \\
        \# Test Cases Mean $\pm$ Std. Dev. & $104.08 \pm 70.42$ \\
        \# Test Cases Range (Min -- Max) & $6$ -- $1440$ \\
        \# Sparse Examples ($\le 10$ Tests) & $804$ ($10.81\%$) \\
        \bottomrule
    \end{tabular}
    \caption{Dataset statistics of TACO subset derived from \cite{luo2025deepcoder}. ``Sparse Examples'' denotes problems with limited unit tests ($\le 10$ inputs).}\label{tab:unit_test_stats}
\end{table}

\section{Additional Main Results}
\label{sec:appendix_qwen3_4b_main_results}
To further examine the robustness of VeRPO under a smaller backbone, we additionally conduct the main experiment with Qwen3-4B. As shown in Table~\ref{tab:qwen3_4b_main_results}, VeRPO consistently outperforms GRPO across all evaluated benchmarks, yielding an average absolute improvement of +3.53 points in pass@1. These results provide additional evidence that the benefit of VeRPO is not specific to the Qwen3-8B backbone used in the main experiments.

\section{Dataset Details}\label{appendix_dataset_detail}

We utilize a filtered subset of the TACO dataset derived from \cite{luo2025deepcoder}. The dataset comprises $7436$ algorithmic problems, each associated with a varying number of unit tests. The statistical distribution of these test cases is summarized in Table~\ref{tab:unit_test_stats}.

\end{document}